\documentclass[lettersize,journal]{IEEEtran}
\usepackage{amsmath,amsfonts}
\usepackage{algorithmic}
\usepackage{algorithm}
\usepackage{array}
\usepackage[caption=false,font=normalsize,labelfont=sf,textfont=sf]{subfig}
\usepackage{textcomp}
\usepackage{stfloats}
\usepackage{url}
\usepackage{verbatim}
\usepackage{graphicx}
\usepackage{cite}
\usepackage{graphics}   
\usepackage{epsfig}     
\usepackage{mathptmx}   
\usepackage{times}      
\usepackage{amssymb}    
\usepackage{booktabs}   
\usepackage{multirow}   
\usepackage{footnote}
\usepackage{color}
\usepackage{hyperref}

\usepackage{float}
\usepackage{enumerate}
\usepackage{bigstrut}
\usepackage{stmaryrd}
\usepackage{gensymb}
\usepackage[scaled=1.0]{helvet}
\usepackage{tabularx}
\usepackage{makecell}
\usepackage{threeparttable}
\usepackage{microtype}

\begin{document}

\title{GESS: Multi-cue Guided Local Feature Learning via Geometric and Semantic Synergy}

\author{
    Yang Yi, Xieyuanli Chen, Jinpu Zhang$^*$, Hui Shen$^*$, and Dewen Hu
    \thanks{
        Y. Yi, X. Chen, J. Zhang, H. Shen, and D. Hu are with the College of Intelligence Science and Technology, National University of Defense Technology, China.
    }
    \thanks{
        * indicates corresponding authors: J. Zhang(zhangjinpu@nudt.edu.cn) and H. Shen(shenhui@nudt.edu.cn)
    }
    \thanks{
        This work has partially been funded by Research on Spatial Perception and Computing Technologies for Dynamic Complex Scenarios Based on Heterogeneous Brain-inspired Intelligence (U25B2069), Autonomous Learning Theory and Applications of Intelligent Unmanned Systems (T2521006) and Multi-modal Object Tracking in Complex Dynamic Scenarios (62506377).
    }
}


\maketitle
\begin{abstract}
Robust local feature detection and description are foundational tasks in computer vision. Existing methods primarily rely on single appearance cues for modeling, leading to unstable keypoints and insufficient descriptor discriminability. In this paper, we propose a multi-cue guided local feature learning framework that leverages semantic and geometric cues to synergistically enhance detection robustness and descriptor discriminability.  Specifically, we construct a joint semantic-normal prediction head and a depth stability prediction head atop a lightweight backbone. The former leverages a shared 3D vector field to deeply couple semantic and normal cues, thereby resolving optimization interference from heterogeneous inconsistencies. The latter quantifies the reliability of local regions from a geometric consistency perspective,  providing deterministic guidance for robust keypoint selection. Based on these predictions, we introduce the Semantic-Depth Aware Keypoint (SDAK) mechanism for feature detection. By coupling semantic reliability with depth stability, SDAK reweights keypoint responses to  suppress spurious features in unreliable regions. For descriptor construction, we design a Unified Triple-Cue Fusion (UTCF) module, which employs a semantic-scheduled gating mechanism to adaptively inject multi-attribute features, improving descriptor discriminability. Extensive experiments on four benchmarks validate the effectiveness of the proposed framework. The source code and pre-trained model will be available at: https://github.com/yiyscut/GESS.git.
\end{abstract}


\section{Introduction}

\begin{figure}[htbp]
    \centering
    \includegraphics[width=0.48\textwidth]{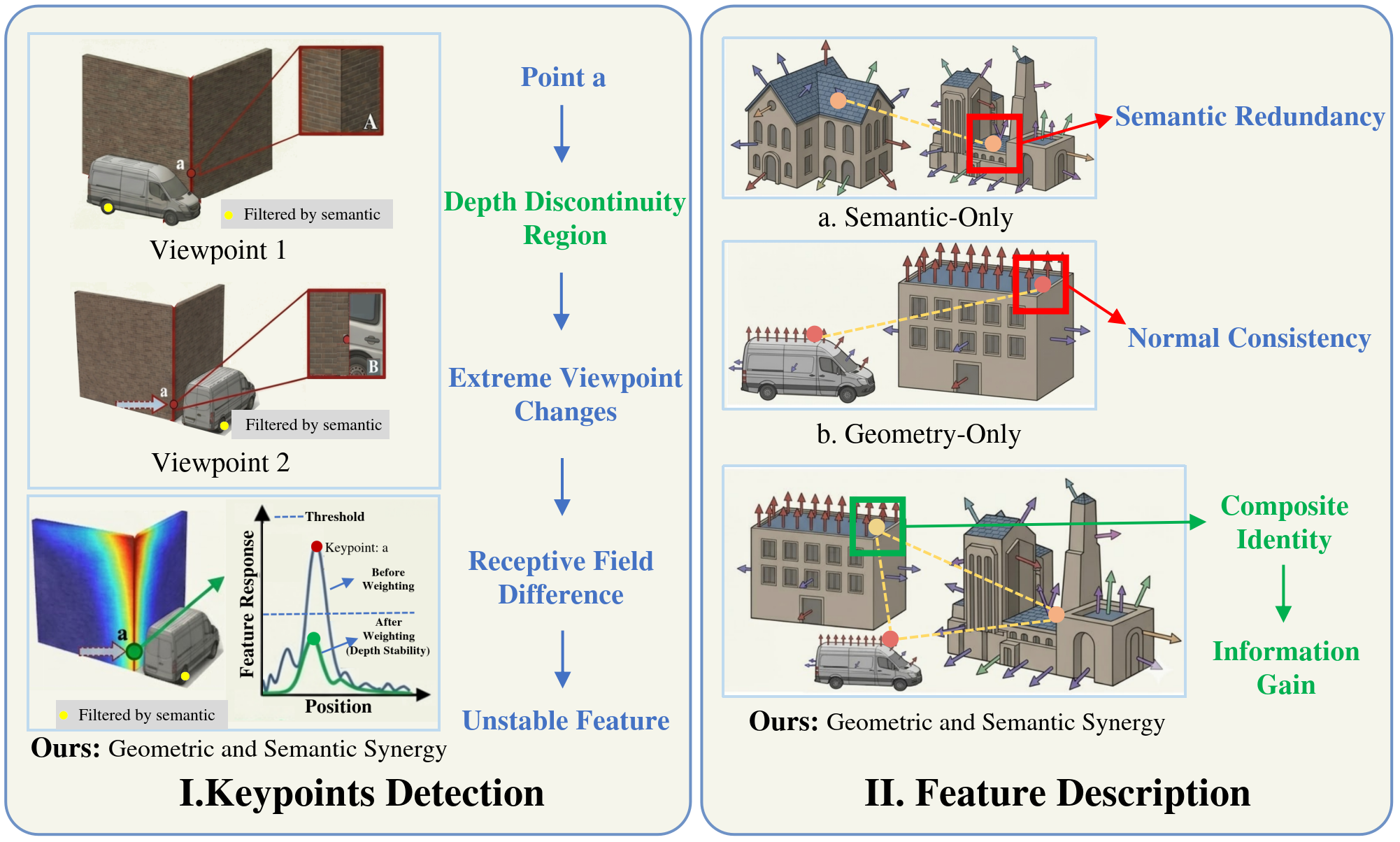}
    \caption{Schematic of geometric and semantic synergy for keypoint detection and feature description. Left: Re-weighting by incorporating depth discontinuity regions and semantic confidence to synergistically enhance the robustness of keypoint detection. Right: Constructing composite identity features to boost information gain, empowering descriptors with superior discriminative performance.}
    \label{introduction_pic}
\end{figure}

Local feature detection and description serve as the core foundations for tasks such as image matching\cite{hpatches,1008392}, visual localization\cite{aachen}, 3D reconstruction\cite{5226635,semantic2018}, and SLAM\cite{MurArtal2017, 10851359}. A reliable local feature representation necessitates both high repeatability and discriminability\cite{r2d2}. Thus, keypoints should remain stable across variations in viewpoint, scale, and illumination, while descriptors must be distinctive enough to support accurate correspondence estimation. In recent years, deep-learning-based approaches have achieved substantial performance gains over traditional handcrafted features\cite{lowe2004distinctive,rublee2011orb,bay2006surf,8848801}  in complex environments. However, most existing methods\cite{superpoint,aslfeat,d2net,disk,zhao2022alike,wang2023attention,8630864,11373611} still rely heavily on a single-cue modeling paradigm driven by visual appearance. In the presence of extreme viewpoint variations or texture-less scenes, relying solely on superficial texture information makes it difficult to maintain stable feature representations, frequently resulting in keypoint drift and mismatches.

To overcome the limitations of relying solely on texture information, an emerging trend involves leveraging auxiliary cue fusion mechanisms to bolster feature discriminability. While some methods\cite{xue2023sfd2, MENG2025129567, 10812856, wu2025segment, 10741542} attempt to incorporate semantic priors alongside texture, naive semantic modeling often suffers from ``semantic redundancy'', which leads to a degradation in discriminative power. Specifically, in scenes containing many homogeneous objects (e.g., building complexes), category similarity provides little additional discriminative information. This results in the descriptor's limited ability to differentiate intra-class fine-grained structures (see Fig. \ref{introduction_pic}.(II.a)). Conversely, another class of methods\cite{liu2025liftfeat} introduces geometric priors but frequently encounters perceptual confusion when facing geometric homogeneity. When multiple surfaces share similar normal distributions, such as building facades or vehicle side panels, normal cues provide limited additional discriminative information. (see Fig. \ref{introduction_pic}.(II.b)).

Inspired by the human visual system\cite{ernst2002humans}, humans perceiving complex scenes not only utilize surface-level color and texture but also instinctively integrate multiple cues, such as semantics (e.g., object categories) and geometric space, to precisely establish correspondences. However, high-level semantics and low-level geometric features are separated by a significant abstraction gap and are prone to divergent failure modes—specifically the semantic redundancy and normal consistency issues discussed above. Constrained by such inherent representational inconsistency, traditional fusion paradigms often suffer from optimization conflicts. Such conflicts cause features across different levels to interfere with each other during gradient updates, making it difficult for the model to reach an effective equilibrium in representation.\cite{8578879}.

To address this challenge, we design a joint semantic-normal prediction head. This module achieves physical alignment of heterogeneous information by constructing a 3D vector field. Under this mechanism, semantics and surface normals are no longer predicted via independent branches. Instead, they are unified into a spatial vector representation. The vector orientation defines the surface normal, while its magnitude quantifies semantic saliency. 
Consequently, this novel joint representation endows discrete semantic cues with explicit physical significance. This coupling ensures that the gradient flow simultaneously captures categorical attributes and geometric details, effectively resolving the gradient conflicts prevalent in the independent optimization of heterogeneous features.
Building upon this, to address the aforementioned inherent failure modes of semantic and normal cues in specific scenarios, we further design the Unified Triple-Cue Fusion (UTCF) module. This module introduces a semantic-scheduled gating mechanism that dynamically modulates deep injection of visual texture, surface normal, and semantic features based on local scene characteristics. Consequently, it constructs collaborative representations with a ``composite identity'' for each feature point (as shown in the green box of Fig. \ref{introduction_pic}.(II)). This hierarchical modeling—extending from low-level gradient alignment to high-level feature adaptive fusion—ensures that the feature descriptors maintain exceptionally high discriminativity even in complex environments.

Meanwhile, recognizing that highly discriminative descriptors must be rooted in highly reliable keypoints, we further introduce a depth stability prediction head and a Semantic-Depth Aware Keypoints (SDAK) enhancement mechanism. 
As illustrated in Fig. \ref{introduction_pic}.(I), this mechanism optimizes the keypoint distribution via dual constraints. Beyond leveraging semantic information to filter out category-unstable features (e.g., the yellow dots on the vehicle), it specifically addresses geometric failures at depth boundaries (e.g., Point a). Such boundary points are highly prone to matching collapse under extreme viewpoint changes due to the significant changes of their receptive fields. To this end, SDAK utilizes the predicted geometric stability map to dynamically re-weight the original response map (blue curve), suppressing the response significantly below the threshold (green curve) to effectively filter out these geometrically unreliable regions.
This strategy ensures that highly discriminative descriptors are supported by reliable keypoints, significantly enhancing the versatility of features across various extreme degraded scenarios.

Finally, to consolidate the representational foundation for multi-source cue collaborative modeling, we introduce a multi-scale fusion mechanism atop the backbone network\cite{wang2025lsnetlargefocussmall}. This mechanism is designed to generate multi-dimensional discriminative features that possess both global semantic consistency and local geometric precision. The main contributions of this paper are as follows:

(i) We propose a multi-cue guided local feature learning framework that jointly exploits semantic saliency, surface normal, and depth stability to synergistically enhance both keypoint detection and descriptor learning.

(ii) We design a Semantic-Normal Coupling Prediction Head and a UTCF module, which adaptively integrate visual texture, surface normal, and semantic features via a semantic-scheduled gating mechanism to boost descriptor discriminability.

(iii) To improve detection robustness, we develop a depth stability prediction head and a SDAK mechanism that dynamically reweight keypoint responses based on depth discontinuities and semantic priors, effectively filtering out unstable candidates.

(iv) Extensive evaluations demonstrate that our method achieves superior performance across diverse scenarios. Specifically, it excels in image matching, relative pose estimation, and 3D reconstruction, while maintaining competitive results in visual localization benchmarks.

    
\section{Related Works}
\subsection{Local Feature Detection and Description}
Traditional local feature methods primarily rely on hand-crafted geometric criteria or statistical heuristics. Representatively, SIFT\cite{lowe2004distinctive} introduced a scale-space-based keypoint detection and gradient histogram description strategy, achieving scale and rotation invariance and establishing itself as a classic benchmark. Subsequently, ORB\cite{rublee2011orb} enhanced algorithmic robustness for engineering applications by optimizing FAST\cite{4674368} corner detection and the BRIEF\cite{10.1007/978-3-642-15561-1_56} descriptor while maintaining real-time performance, which led to its widespread adoption in resource-constrained visual tasks.

With the rise of deep learning, local feature processing has transitioned toward an end-to-end learning paradigm. Early works, such as LIFT\cite{yi2016liftlearnedinvariantfeature} and L2-Net\cite{8100132}, achieved the end-to-end reconstruction of feature extraction pipelines and the learning of highly discriminative descriptors, respectively. Subsequently, SuperPoint\cite{detone2018superpoint} introduced a self-supervised strategy, pioneering an efficient joint learning framework for both interest points and descriptors. To further address the inherent limitations of foundational models in stability and generalization, subsequent research has pursued multi-dimensional improvements: firstly, manual priors, deformable convolutions, and reliability estimation mechanisms\cite{barrosolaguna2019keynetkeypointdetectionhandcrafted, aslfeat, r2d2} were incorporated to effectively enhance robustness in complex environments; secondly, context-aware modules\cite{wang2022mtldesc, luo2019contextdesc} and domain adaptation techniques\cite{10144412} were leveraged to expand local receptive fields and boost cross-domain representation; finally, to satisfy real-time requirements, probabilistic feature sampling and differentiable detection frameworks\cite{disk, zhao2022alike} were proposed, achieving a favorable trade-off between speed and accuracy. Collectively, these advancements have significantly broadened the practical applicability of local feature learning systems.

While these methods significantly enhance scene robustness, most paradigms remain constrained by appearance-based single-cue modeling\cite{10485434}, lacking a high-level understanding of semantic logic and explicit constraints on micro-spatial geometric structures. This inevitably limits the discriminative ceiling and matching robustness of features in extreme environments.

\subsection{Semantic-Guided Local Feature Learning}
To address the performance degradation of local features in long-term and drastically changing scenes, researchers have recently advocated for the deep coupling of semantic information as a core cue within the feature learning paradigm. Specifically, SFD2\cite{xue2023sfd2} proposed an implicit feature guidance mechanism that effectively transfers semantic knowledge into local feature representations by introducing feature consistency losses in the intermediate layers of the encoder. Similarly, Wang et al.\cite{10741542} integrated semantic cues into a reinforcement learning reward mechanism, achieving adaptive representation enhancement in complex and volatile environments through task-driven optimization. With the advent of foundation models, SAMFeat\cite{wu2025segment} leveraged the generalization power of the Segment Anything Model(SAM)\cite{kirillov2023segany} to assist feature extraction, yielding substantial gains in matching performance. However, due to the class-agnostic nature of SAM, it often struggles to provide precise semantic constraints when facing complex dynamic interference. In addition, concurrent works such as SPADesc\cite{MENG2025129567}, SAGA-Feat\cite{10.1016/j.neucom.2025.131349}, and SDE2D\cite{10812856} have explored semantic-guided feature learning from various dimensions, including global semantic correlation, spatial context aggregation, and weak-textured region identification.

While these methods eliminate the need for extra semantic labels during the inference phase and strike a favorable balance between efficiency and accuracy, achieving the deep fusion of semantic and texture features in highly dynamic scenes remains a formidable challenge.

\subsection{Geometry-Guided Local Feature Learning}

\begin{figure*}[htbp]
    \centering
    \includegraphics[width=1\textwidth]{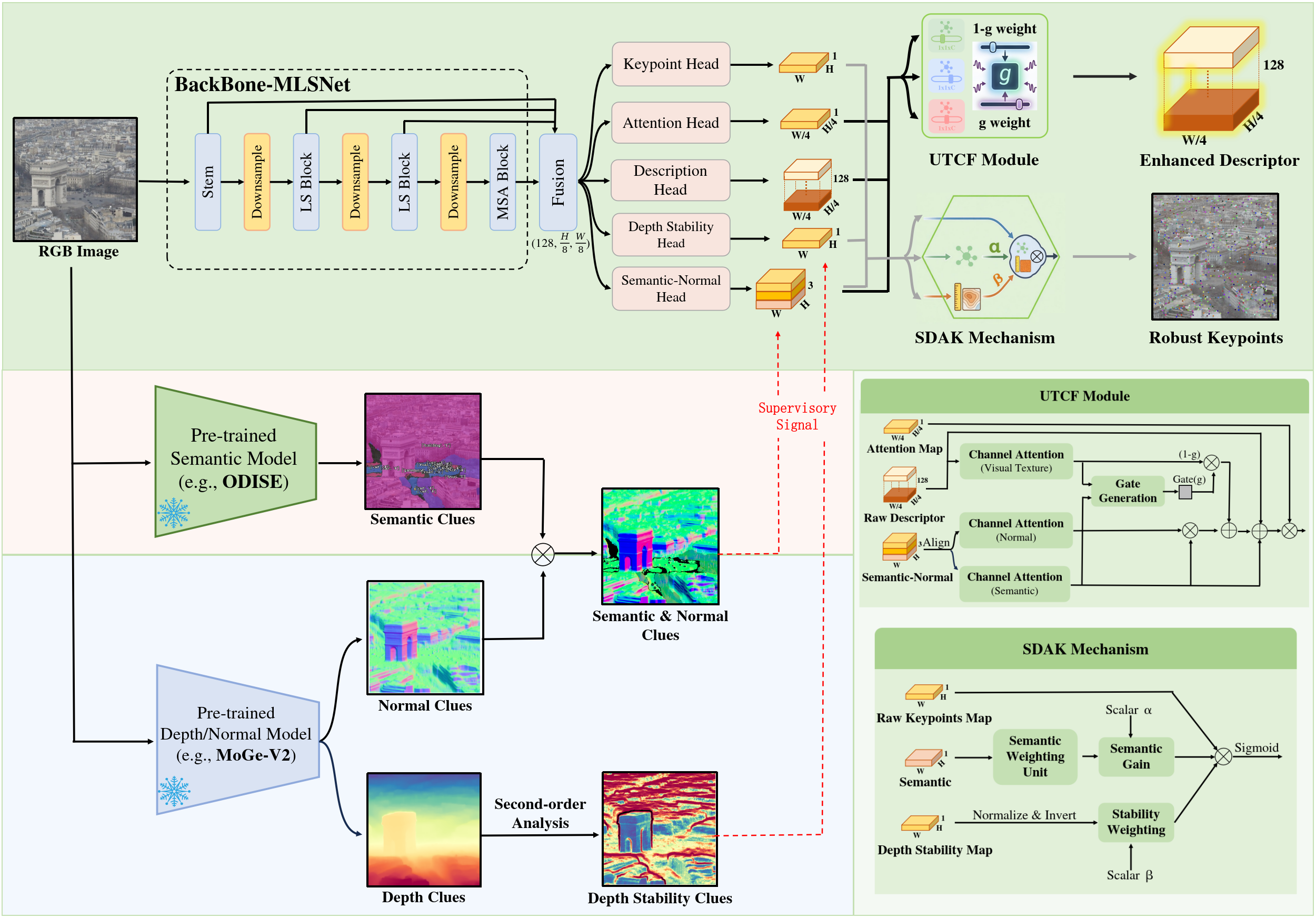} 
    \caption{Overview of the proposed framework. Built upon the multi-scale feature extraction backbone (MLSNet), our framework predicts multi-dimensional cues—including semantics, normals, and depth stability—through multi-branch heads. The UTCF module then adaptively integrates visual texture, normal, and semantic features to significantly enhance descriptor discriminability. Meanwhile, the SDAK mechanism dynamically reweights the initial heatmap, guiding the model to select keypoints that are semantically meaningful and robustly localized.}
    \label{model}
\end{figure*}

Beyond leveraging environmentally invariant semantic information, the geometric structure of a scene plays an indispensable role in providing viewpoint invariance and scale consistency. To this end, researchers have further explored incorporating geometric priors to guide the construction of local feature descriptors, aiming for high-precision, geometrically robust representations.

Regarding cross-view geometric correlations, PoSFeat\cite{li2022decoupling} adheres to the ``describe-then-detect'' decoupling paradigm and innovatively employs epipolar constraints as geometric supervision signals to guide the network in learning cross-view consistent geometric representations. At the multi-scale representation level, MTLDesc\cite{wang2022mtldesc} strictly follows scale-space theory to construct a multi-scale feature pyramid detection module, effectively enhancing the localization stability of keypoints across varying resolutions. Distinct from the aforementioned projection-based or multi-resolution strategies, an alternative trajectory involves directly incorporating intrinsic physical geometric attributes, such as surface normals, to provide finer spatial consistency constraints. As a representative method in this direction, LiftFeat\cite{liu2025liftfeat} integrates 3D surface normal information into the feature encoding process, significantly strengthening the descriptors' robustness against surface pose variations. However, the inherent geometric homogeneity of local features limits the generalizability of these methods in complex scenarios.

In summary, existing research has validated the respective efficacy of semantic priors in identifying long-term stable regions and geometric cues in reinforcing spatial consistency. However, due to the lack of deep modeling regarding the intrinsic correlation between semantic logic and geometric structures, existing paradigms still struggle to fundamentally circumvent the aforementioned failure modes. Consequently, features fail to maintain robust representational consistency when confronted with complex scenes. To address these challenges, this paper proposes a multi-cue collaboratively guided framework for local feature learning. By integrating semantic, normal, and depth reliability cues, the framework establishes a synergistic constraint mechanism among heterogeneous cues, further enhancing the scene robustness of feature detection and elevating the discriminative upper bound of descriptors.

\section{METHODOLOGY}
\subsection{Overview}
This chapter elaborates on the proposed multi-cue guided local feature learning framework, which integrates semantic saliency, surface normals, and spatial geometric stability to comprehensively enhance keypoint detection robustness and descriptor discriminativeness. At the architectural level, , as illustrated in Fig. \ref{model}, the model first employs a multi-scale backbone (MLSNet) to capture multi-level feature representations that balance global semantic consistency with local geometric precision. In the prediction stage, we construct a multi-branch parallel task-specific architecture. Beyond the standard descriptor, keypoints, and attention branches, we introduce two auxiliary heads for semantic-normal (See III.B) and depth stability  (See III.C) prediction. These tasks are optimized in an end-to-end manner, supervised by pseudo-labels generated from high-performance pre-trained models. Based on these predicted multi-dimensional cues, the UTCF module (See III.D) achieves a decoupled fusion of descriptors and semantic-normal cues to bolster the descriptor's discriminative power. Simultaneously, the SDAK mechanism (See III.E) performs a re-weighting modulation on the keypoint heatmap by integrating semantic gain and geometric stability weights. This guides the model to filter and select feature points with clearer physical significance and more robust localization.

\subsection{Semantic-Normal Head}
Departing from conventional independent branch prediction, we introduce a coupled prediction head. This design enforces the network to learn the intrinsic consistency between physical geometry and categorical attributes, ensuring that the generated features possess both robust local geometry and macroscopic semantic constraints. This module predicts a three-dimensional vector field $\vec{V} \in \mathbb{R}^{H \times W \times 3}$. It decouples the output at each pixel location into geometric and attributive dimensions: the normalized vector direction $\vec{n} = \frac{\vec{V}}{\|\vec{V}\|_2}$ is used to extract the surface normal representing local topology, while the vector magnitude $s = \|\vec{V}\|_2$ is employed to characterize the semantic saliency of the region. The core advantage of this design lies in the orthogonality of gradient directions and the stability of optimization. We define the joint loss function (with the weight of each term set to 1) as follows:
\begin{equation}
{{\mathcal{L}}_{\mathrm{couple}}} = {{\mathcal{L}}_{\mathrm{seg}}}(s, s^*) + {{\mathcal{L}}_{\mathrm{normal}}}(\vec{n}, \vec{n}^*)
\end{equation}
where $s^*$ and $\vec{n}^*$ are the pseudo ground-truth semantic category and surface normal, respectively. According to the chain rule, the gradient of the total loss with respect to the output vector field $\vec{V}$ can be expanded as:
\begin{equation}
\frac{\partial {{\mathcal{L}}_{\mathrm{couple}}}}{\partial \vec{V}} = {\frac{\partial {{\mathcal{L}}_{\mathrm{seg}}}}{\partial s} \vec{n}}+ {\frac{1}{s} (\mathbf{I} - \vec{n}\vec{n}^T) \frac{\partial {{\mathcal{L}}_{\mathrm{normal}}}}{\partial \vec{n}}}
\end{equation}
This coupling mechanism decomposes the gradient flow into mutually orthogonal radial (semantic) and tangential (surface normal) components. This decomposition helps reduce gradient interference 
between semantic and geometric supervision, while leveraging the mathematical complementarity between magnitude and direction to achieve adaptive gradient compensation. Consequently, it eliminates representation conflicts among heterogeneous features and effectively corrects the gradient estimation biases that may arise from task decoupling.

To optimize the aforementioned coupled prediction head and facilitate the learning of semantic-aware geometric features, we design a joint semantic and geometric supervision strategy. This strategy constructs high-quality supervision targets for the direction and magnitude of the vector field $\vec{V}$, respectively. Specifically, following the semantic partitioning criterion of SFD2\cite{xue2023sfd2}, we classify the fine-grained semantics extracted by the ODISE\cite{xu2023open} model into four categories, and assign corresponding reliability weights to generate a pixel-wise semantic weight map. Meanwhile, high-precision surface normals produced by MoGe-V2\cite{wang2025moge} are employed as the geometric benchmark. Finally, to accelerate training, we fuse semantic weights with surface normal maps pixel-wise to generate semantic-normal vector labels. This supervisory signal constrains the network to focus on stable geometric structures while suppressing dynamic noise, thereby learning more discriminative feature representations in complex environments.

Guided by the above supervision strategy, we define the following objective function to jointly optimize the geometric and semantic attributes of the coupling prediction head.

\textbf{Normal Constraint.} We employ an angle-based normal loss to directly constrain the angular deviation between the predicted normal $\vec{n}$ and the pseudo ground truth normal $\vec{n}^*$ generated by MoGe-V2\cite{wang2025moge}:
\begin{equation}
{{\mathcal{L}}_{\mathrm{normal}}}=\frac{1}{|\mathrm{\Omega}|}\sum_{(i,j) \in \mathrm{\Omega}}{\arccos{(}}\vec{n}(i,j)\cdot{{\vec{n}}^{*}}(i,j))
\end{equation}
where $\Omega$ denotes the set of pixel positions involved in the calculation, and similarly hereinafter.

\textbf{Semantic Constraint.} For the semantic prediction task, the vector magnitude $s = \|\vec{V}\|_2$ is fed into a lightweight classifier to obtain the predicted probability $p_m(i, j)$ of each pixel belonging to class $m$. Subsequently, in conjunction with the semantic ground truth labels $y_m(i, j)$ generated by ODISE\cite{xu2023open}, the model is supervised using the following cross-entropy loss function:
\begin{equation}
{{\mathcal{L}}_{\mathrm{seg}}}=-\frac{1}{|\mathrm{\Omega}|}\sum_{(i,j) \in \mathrm{\Omega}}\sum_{m=1}^{M} y_{m}(i,j)\cdot\log p_{m}(i,j)
\end{equation}
where $M$ denotes the total number of semantic categories, which is 4.

\subsection{Depth Stability Head}
To quantify the 3D structural stability of a scene and suppress depth noise interference, we introduce the Depth Stability Head. This module takes high-resolution shallow features from the backbone as input, aiming to leverage their superior spatial fidelity to preserve primitive local gradient details, thereby providing a fundamental basis for accurately determining micro-structural stability. Specifically, this branch utilizes a lightweight geometric perceptron (consisting of two $3 \times 3$ convolutional layers) to perform non-linear refinement on the information within the shallow features. Subsequently, a sigmoid activation function is employed to map the features into a $[0, 1]$ continuous confidence space, explicitly modeling the scene's geometric reliability. By acquiring critical geometric constraint priors with minimal overhead, this design effectively ensures superior repeatability and stability of the features in 3D space.

Following the architectural definition of the Depth Stability Head, we establish its corresponding physical supervision logic. First, Gaussian smoothing is applied to the dense depth maps generated by MoGe-V2\cite{wang2025moge} to mitigate sampling noise. Subsequently, the Sobel operator is employed to extract the first-order gradient magnitude $\delta$, facilitating the precise localization of depth discontinuities within the scene. Simultaneously, the Laplacian operator is utilized to compute the second-order response $L$, serving as a proxy indicator for surface curvature. Finally, regions with pronounced geometric transitions are mapped to stability weights $s^*$ via an exponential decay function, which is formally defined as follows:
\begin{equation}
    s^* = \epsilon + (1 - \epsilon) \cdot \exp(-\gamma \cdot (\alpha_\delta \delta + \alpha_l L))
\end{equation}
where the weighting coefficients are empirically set as $\alpha_\delta = 2.0$ and $\alpha_l = 1.0$, the decay intensity $\gamma = 3.0$, and the floor threshold $\epsilon = 0.2$. The essence of this design lies in quantifying the geometric instability triggered by depth discontinuities, thereby establishing a fundamental criterion for assessing the physical reliability of the scene structure.

Finally, we introduce the Geometric Stability Loss to explicitly supervise the predicted map $s$. Formulated based on the $L_1$ distance, this loss function is designed to guide the model in capturing geometric stability priors:
\begin{equation}
\mathrm{{{\mathcal{L}}_{sta}}=\frac{1}{|\Omega|}\sum_{(i,j) \in \Omega}{\|}s(i,j)-s^{*}(i,j){\|}_{1}}
\end{equation}
By minimizing this distance constraint, the model learns to effectively characterize the spatial distribution of geometric stability, providing a reliable prior for subsequent modules to identify and suppress unstable regions.

\subsection{Unified Triple-Cue Fusion Module}
To generate robust and highly discriminative descriptors under large viewpoint changes, illumination variations, and low-texture scenarios, we design the UTCF module. This module introduces normal geometry and semantic priors to achieve complementary enhancement with texture information. The overall pipeline consists of three core stages: Cue-Specific Channel Calibration, Semantic-Modulated Gating Fusion, and Feature Refinement with Residual Connection.

\subsubsection{Cue-Specific Channel Calibration}
Before fusing multiple cues, it is essential to account for their inherent disparities in information density, feature distribution, and physical semantics. To this end, we perform cue-specific channel calibration, which independently recalibrates the texture (initial visual descriptors), normal, and semantic cues. This process aims to emphasize discriminative intra-modal channels and suppress redundant information, thereby providing purified inputs for subsequent fusion. Specifically, normal features are first passed through a projection layer—comprising a $1 \times 1$ convolution, Batch Normalization, and ReLU—to align their channel dimensions with those of the descriptors. Meanwhile, semantic cues undergo two $3 \times 3$ convolutional layers to extract high-level semantic features with an output dimensionality of $48$. Subsequently, independent channel attention mechanisms are applied to these three branches. For a given cue feature $F_c \in \mathbb{R}^{H \times W \times C}$ (where $c \in \{t, n, s\}$ denotes texture, normal, and semantic cues, respectively), the corresponding channel attention weights are generated as follows:
\begin{equation}
    W_c = \sigma\left( \text{MLP}_c \left( \text{GAP}(F_c) \right) \right)
\end{equation}
Specifically, $\text{GAP}$ denotes Global Average Pooling, which aims to squeeze the spatial dimensions into a $1 \times 1$ channel descriptor to aggregate global context. $\text{MLP}_c$ utilizes a reduction ratio to construct a low-overhead two-layer perceptron, designed to strengthen non-linear interactions between channels. Finally, $\sigma$ represents the Sigmoid function, which maps the output to the $(0, 1)$ interval to generate channel-wise scaling weights. Ultimately, the calibrated features are defined as:
\begin{equation}
    F'_c = W_c \odot F_c
\end{equation}
where $\odot$ denotes the Hadamard product.

The MLP parameters for each cue are entirely independent, which decouples the feature responses of physically distinct signals and prevents mutual cross-talk. This enables the network to learn specialized attention patterns tailored to the unique statistical characteristics of each cue.

\subsubsection{Semantic-Modulated Gating Fusion}
The core of the UTCF module is a dynamic gating mechanism based on semantic priors, which acts as a ``dynamic arbitrator'' among different cues. We first concatenate the channel-calibrated texture features $F'_t$, normal features $F'_n$, and the lightweight semantic gating features $G_s$ along the channel dimension, and feed them into a lightweight convolutional mapping network $\Phi$ to generate a spatially adaptive gating weight map $g \in [0,1]^{H \times W}$:
\begin{equation}
    g = \sigma(\Phi([F'_t; F'_n; F'_s]))
\end{equation}
where $[\cdot]$ denotes channel concatenation, and $\Phi$ consists of two 1×1 convolutions: the first reduces the channel dimension of the concatenated features to 128, and the second further compresses it to a single channel, followed by a Sigmoid function $\sigma$ for activation. This gating weight $g$ dynamically adjusts the contribution ratio between texture and normal features, and the fused intermediate feature $F_{\text{fused}}$ is computed as:
\begin{equation}
    F_{fused} = (1 - g) \odot F'_t + g \odot F'_n
\end{equation}

The core idea of this design is to leverage semantic priors to ``modulat'' the combination of visual texture and geometric information: in regions where semantic priors indicate low texture, the gating weight $g$ tends to increase, making the model more reliant on the stability of geometric normals; whereas in semantically complex or dynamic regions, the gate suppresses the normal contribution and highlights the discriminative power of texture descriptors. This mechanism ensures that the descriptor maintains high discriminative power across various extreme scenarios through the dynamic complementarity of multiple cues.

\subsubsection{Feature Refinement and Residual Connection}
To integrate macro-semantic priors while preserving fine-grained texture details, we propose a feature refinement strategy transitioning from ``local infusion'' to ``global calibration''. Specifically, a semantic increment $D_s$ is first extracted from the calibrated semantic features $F'_s$, with its dimensionality aligned to the gated fusion features $F_{fused}$ via a $1 \times 1$ convolutional layer. Subsequently, this semantic increment is injected into the preliminary fusion result to yield the coarse refined feature:
\begin{equation}
    F_{refine} = F_{fused} + \mu \cdot D_s
\end{equation}
where $\mu = 0.1$ is a scaling factor used to modulate the intensity of semantic intervention, preventing the over-smoothing effect of semantic features from obscuring low-level texture details. This feature is further processed through an output projection layer for non-linear transformation, resulting in the refined feature $F_{refine}$. Finally, to ensure the network output maintains the strong discriminative power of the initial texture, we introduce a residual connection mechanism that integrates the refined feature with the initial texture descriptor $D_{initial}$. The final augmented descriptor $D_{output}$ is produced via recalibration with the attention map $W_{map}$:
\begin{equation}
    D_{output} = W_{map} \odot (F_{refine} + D_{initial})
\end{equation}

This design ensures that the descriptor captures high-frequency texture information while effectively absorbing macro-semantic consistency and geometric constraints from normal cues. By incorporating physical consistency constraints, the entire fusion strategy extends the representation from a purely appearance-based dimension to a multi-dimensional latent space synchronized with geometry and semantics, significantly enhancing the discriminative power of the descriptor in extreme environments.

\subsection{Semantic-Depth Aware Keypoints Detection}
Upon obtaining the initial keypoint heatmap $K_{map}$, we propose the SDAK mechanism to filter out unstable features in complex 3D dynamic scenes. The core intuition lies in formulating macro-semantic stability and micro-geometric reliability as dual spatial constraints to modulate the feature response. While texture-only detectors\cite{superpoint, r2d2, wang2022mtldesc} are prone to texture-induced false positives at occlusion boundaries or on dynamic objects, SDAK performs spatial arbitration over keypoint distributions by leveraging cross-dimensional priors. Specifically, the mechanism integrates $K_{map}$ with the semantic confidence map $S_{map}$ and the depth reliability map $R_{map}$. First, the semantic input is processed through a lightweight convolutional network to generate a saliency mask $S_{mask} = \sigma(\text{Conv}(S_{map})) \in [0,1]$, identifying regions with long-term semantic consistency. Simultaneously, the depth reliability map is resampled to align with the heatmap dimensions to perceive and suppress geometrically degraded regions. The final refined heatmap is formulated as:
\begin{equation}
    K_{final} = K_{map} \odot (1 + \alpha \cdot S_{mask} + \beta \cdot R_{map})
\end{equation}
where $\alpha, \beta$ are learnable scalars that modulate the gain of semantic enhancement and the weight of geometric reliability, respectively. Under this multi-modal constraint, the network is guided to suppress regions with geometric uncertainty or semantic instability, thereby extracting keypoints with high repeatability and physical consistency in challenging environments.

\begin{figure*}[htbp]
    \centering
    \includegraphics[width=1\textwidth]{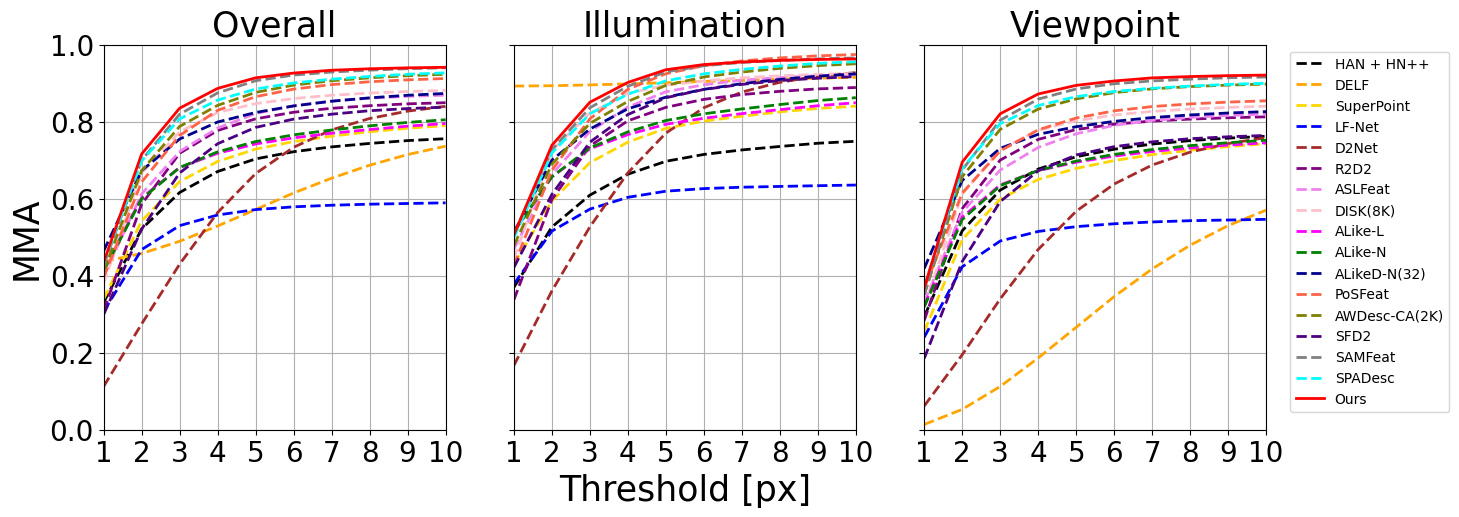} 
    \caption{Comparison of Mean Matching Accuracy (MMA) on the HPatches dataset across varying pixel error thresholds. Our model consistently achieves the best overall performance compared to state-of-the-art methods.}
    \label{hpatch}
\end{figure*}

\begin{figure*}[htbp]
    \centering
    \includegraphics[width=1\textwidth]{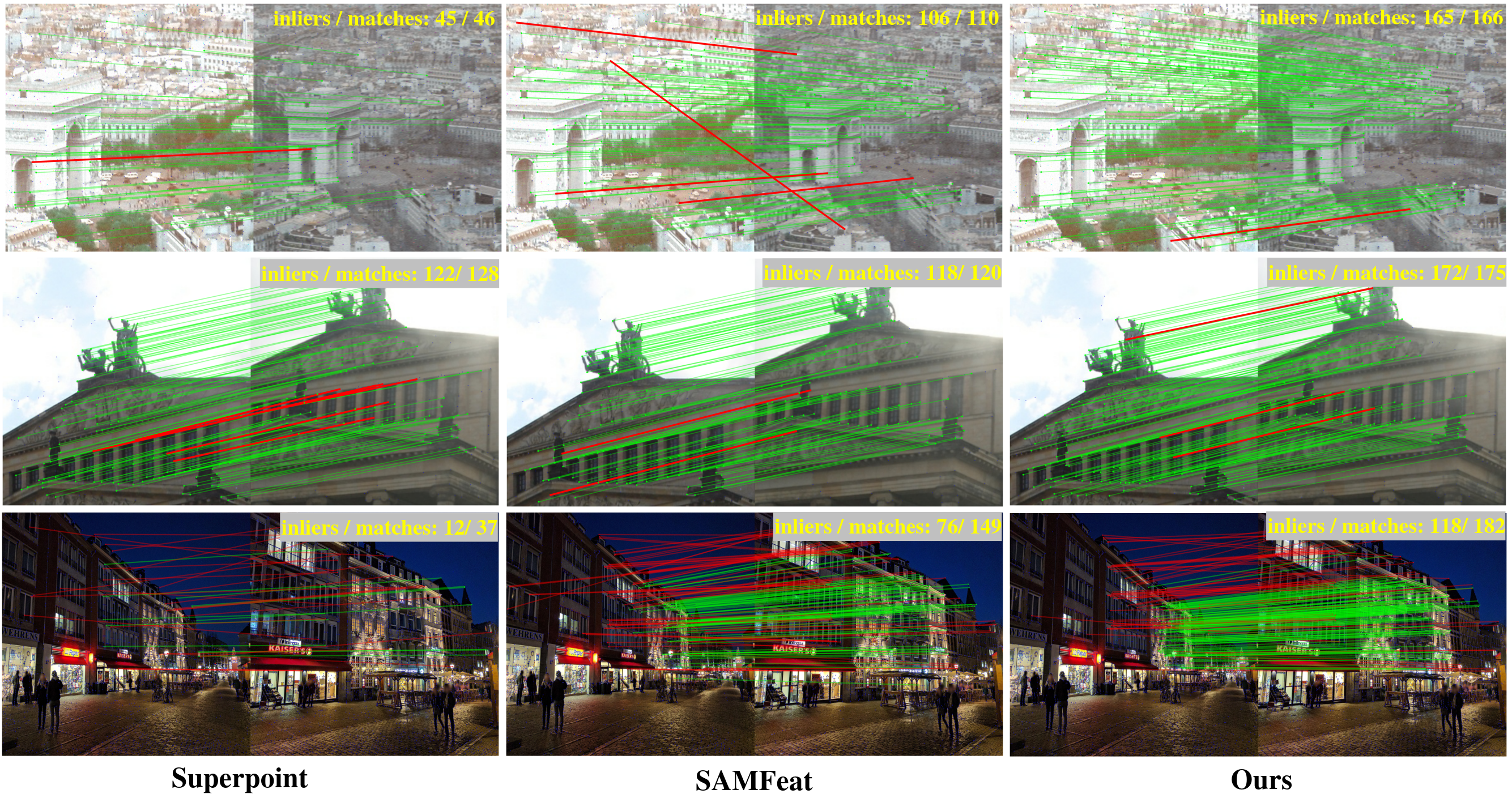}
    \caption{Qualitative comparison of matching performance across various challenging scenarios.  Green lines represent correct inlier matches, while red lines denote outliers.  Our model consistently identifies a higher number of inliers across variations in illumination, low-texture scenes, and viewpoint changes.}
    \label{match}
\end{figure*}

\subsection{Network Training}
\subsubsection{Implementation}
We implemented the proposed framework using the PyTorch framework. During the training phase, to ensure a fair comparison, we strictly followed the experimental protocols established by MTLDesc\cite{wang2022mtldesc} and SAMFeat\cite{wu2025segment}, training exclusively on the MegaDepth sub-dataset\cite{megadepth} without incorporating any additional data resources such as COCO\cite{lin2014coco}. The model was optimized using the Adam optimizer with an initial learning rate of $1 \times 10^{-3}$, a weight decay of $1 \times 10^{-4}$, and a batch size of 8. Regarding the evaluation pipeline, we implemented physical parallelization between the training and evaluation stages to optimize the experimental workflow and expedite the validation cycle. Specifically, while the model training was conducted on an NVIDIA RTX 5090 GPU, all performance benchmarks were executed synchronously on a separate device equipped with an NVIDIA RTX 4090 GPU.

\subsubsection{Joint Loss Function Optimization}
The total training objective of the model comprises the base feature loss and the three aforementioned losses. The total loss function of the system is defined as follows:
\begin{equation}
    \mathcal{L}={{{{{{\mathcal{L}}_{det}}\mathrm{+\mathcal{L}}}_{\mathrm{des}}}+\mathcal{L}}_{\mathrm{normal}}}+{{\mathcal{L}}_{\mathrm{seg}}}+{{\mathcal{L}}_{\mathrm{sta}}}
\end{equation}
where the fundamental terms $\mathcal{L}_{det}$ and $\mathcal{L}_{des}$ are both inherited from the baseline model\cite{wang2022mtldesc}. Specifically, $\mathcal{L}_{det}$ adopts a weighted binary cross-entropy loss for keypoint detection supervision, whereas $\mathcal{L}_{des}$ utilizes an attention-weighted triplet loss for descriptor learning.
    
\section{EXPERIMENTS}
In this section, we conduct a comprehensive evaluation of the proposed algorithm's performance across four key dimensions: image matching, visual localization, relative pose estimation, and sparse 3D reconstruction. Furthermore, we present detailed ablation studies to quantify the individual efficacy of the constituent sub-modules within the architecture. 

It is important to emphasize that, with the exception of the visual localization tasks , all experimental results were obtained and verified under a consistent local hardware environment to ensure a fair comparison.

\subsection{Image Matching}
\subsubsection{Dataset}
    The HPatches dataset\cite{hpatches} is a widely used benchmark for evaluating local feature descriptors. To ensure reliable evaluation, certain unreliable sequences are excluded following\cite{d2net}.

\subsubsection{Metrics}
    Following the evaluation protocol established in \cite{1498756, wu2025segment}, we employ Mean Matching Accuracy (MMA) and the Area Under the Curve (AUC) as our core evaluation metrics, calculated within an error threshold range of 1 to 10 pixels.

\subsubsection{Results}
    As shown in Fig. \ref{hpatch}, our model achieves competitive performance across the entire range of pixel error thresholds (1-10 px), exhibiting consistent matching robustness under both strict and lenient evaluation criteria. Furthermore, its stable performance in the face of illumination and viewpoint variations further validates its strong generalization capability in complex real-world environments. The quantitative comparison results for image matching are presented in the Tab. \ref{tab_hpatches_01}. Our model exhibits superior performance across all evaluation dimensions, securing the top position in all three core metrics.

    Furthermore, we present qualitative image matching results of the proposed method under various challenging scenarios, including illumination variations, similar textures, and large viewpoint variations in night-time scenes (as shown in Fig. \ref{match}). Even when restricted to basic nearest neighbor (NN) matching, our model consistently demonstrates significant performance advantages.Comparative results with SuperPoint\cite{superpoint} and SAMFeat\cite{wu2025segment} reveal that our model identifies a substantially higher number of inliers, outperforming the baseline methods by a considerable margin. This further validates that the feature representations generated through the fusion of geometric and semantic cues possess high discriminative power, thereby ensuring the reliability of visual associations in challenging environments.

    \begin{table}[htbp]
        \centering
        \renewcommand{\arraystretch}{1.2} 
        \setlength{\tabcolsep}{8pt} 
        
        \caption{Quantitative comparison on the HPatches dataset \cite{hpatches}.}
        \label{tab_hpatches_01}
        
        \begin{threeparttable}
            \begin{tabular}{lccc}        
                \toprule
                \textbf{Methods} & \textbf{MMA@3} & \textbf{AUC@2} & \textbf{AUC@5} \\
                \midrule
                SuperPoint \cite{superpoint} {\scriptsize (CVPRW’18)}      & 59.38 & 37.31 & 55.46 \\
                D2-Net \cite{d2net} {\scriptsize (CVPR’19)}              & 43.51 & 19.65 & 41.85 \\
                R2D2 \cite{r2d2} {\scriptsize (NeurIPS’19)}              & 71.83 & 44.87 & 66.06 \\
                DISK \cite{disk} {\scriptsize (NeurIPS’20)}              & 77.45 & 52.02 & 71.92 \\
                ALIKE-N \cite{zhao2022alike} {\scriptsize (TMM’22)}      & 68.10 & 50.62 & 64.57 \\
                ALIKE-L \cite{zhao2022alike} {\scriptsize (TMM’22)}      & 68.10 & 50.87 & 64.50 \\
                SFD2 \cite{xue2023sfd2} {\scriptsize (CVPR’23)}          & 67.94 & 44.06 & 63.28 \\
                MTLDesc \cite{wang2023attention} {\scriptsize (TPAMI’23)} & 78.77 & 54.26 & 73.68 \\
                SAMFeat \cite{wu2025segment} {\scriptsize (TIP’25)}      & \underline{81.96} & \underline{56.35} & \underline{76.53} \\
                \textbf{GESS (Ours)}                                     & \textbf{83.51} & \textbf{57.56} & \textbf{77.81} \\
                \bottomrule
            \end{tabular}
        \end{threeparttable}
    \end{table}

\subsection{Visual Localization}
    \begin{table*}[htbp]
        \centering
        \caption{Visual localization results on Aachen Day-Night V1.0 and V1.1 datasets \cite{aachen}.}
        \setlength{\tabcolsep}{1pt}
        \begin{tabular}{c l  ccc  ccc  ccc  ccc}
        \toprule
        & & \multicolumn{6}{c}{Aachen V1.0} & \multicolumn{6}{c}{Aachen V1.1} \\
        \cmidrule(lr){3-8}\cmidrule(lr){9-14}
        Group & Methods &
        \multicolumn{3}{c}{Day} &
        \multicolumn{3}{c}{Night} &
        \multicolumn{3}{c}{Day} &
        \multicolumn{3}{c}{Night} \\
        \cmidrule(lr){3-5}\cmidrule(lr){6-8}\cmidrule(lr){9-11}\cmidrule(lr){12-14}
        & &
        $(2^\circ,0.25)$ & $(5^\circ,0.5)$ & $(10^\circ,5)$ &
        $(2^\circ,0.25)$ & $(5^\circ,0.5)$ & $(10^\circ,5)$ &
        $(2^\circ,0.25)$ & $(5^\circ,0.5)$ & $(10^\circ,5)$ &
        $(2^\circ,0.25)$ & $(5^\circ,0.5)$ & $(10^\circ,5)$ \\
        \midrule
        \multirow{3}{*}{V}
        & SeLF\cite{fan2022learning} {\scriptsize{(TIP’22)}}               & -- & -- & -- & -- & -- & -- & -- & -- & -- & 75.0 & \underline{86.8} & \underline{97.6} \\
        & LBR\cite{xue2022efficient} {\scriptsize{(CVPR’22)}}               & \textbf{88.3} & \underline{95.6} & \textbf{98.8} & \underline{84.7} & \underline{93.9} & \textbf{100.0} & \textbf{88.3} & \underline{95.6} & \textbf{98.8} & \underline{77.0} & \textbf{92.1} & \textbf{99.5} \\
        & SFD2\cite{xue2023sfd2}  {\scriptsize{(CVPR’23)}}                   & \underline{88.2} & \textbf{96.0} & \underline{98.7} & \textbf{87.8} & \textbf{94.9} & \textbf{100.0} & \underline{88.2} & \textbf{96.0} & \underline{98.7} & \textbf{78.0} & \textbf{92.1} & \textbf{99.5} \\
        \midrule
        \multirow{11}{*}{G}
        & SIFT\cite{lowe2004distinctive} {\scriptsize{(IJCV’04)}}            & 82.8 & 88.1 & 93.1 & 30.6 & 43.9 & 58.2 & 72.2 & 78.4 & 81.7 & 19.4 & 23.0 & 27.2 \\
        & SuperPoint\cite{detone2018superpoint} {\scriptsize{(CVPRW’18)}}    & 80.5 & 87.4 & 94.2 & 42.9 & 62.2 & 76.5 & 87.9 & 93.6 & 96.8 & 70.2 & 84.8 & 93.7 \\
        & D2-Net~\cite{d2net} {\scriptsize{(CVPR’19)}}                        & 84.8 & 92.6 & \textbf{97.5} & \textbf{84.7} & \underline{90.8} & 96.9 & 84.1 & 91.0 & 95.5 & 63.4 & 83.8 & 92.1 \\
        & R2D2~\cite{r2d2} {\scriptsize{(NeurIPS’19)}}                         & -- & -- & -- & 76.5 & \underline{90.8} & \textbf{100.0} & 88.8 & 95.3 & 97.8 & 72.3 & \underline{88.5} & 94.2 \\
        & ASLFeat~\cite{luo2020aslfeat}  {\scriptsize{(CVPR’20)}}           & -- & -- & -- & 81.6 & \underline{90.8} & \textbf{100.0} & 88.0 & 95.4 & 98.2 & 70.7 & 84.3 & 94.2 \\
        & PoSFeat~\cite{li2022decoupling}  {\scriptsize{(CVPR’22)}}          & -- & -- & -- & 81.6 & 87.8 & \textbf{100.0} & -- & -- & -- & 73.8 & 87.4 & \textbf{98.4} \\
        & AWDesc-CA(2K) \cite{wang2023attention} {\scriptsize{(TPAMI’23)}}   & -- & -- & -- & -- & -- & -- & -- & -- & -- & 74.3 & 86.9 & \underline{96.9} \\
        & XFeat~\cite{potje2024cvpr} {\scriptsize{(CVPR’24)}}                & 84.7 & 91.5 & 96.5 & 77.6 & 89.8 & 98.0 & -- & -- & -- & -- & -- & -- \\
        & LiftFeat\cite{liu2025liftfeat} {\scriptsize{(ICRA’25)}}           & \textbf{87.6} & \underline{93.1} & 97.1 & 82.1 & 89.9 & \underline{99.1} & -- & -- & -- & -- & -- & -- \\
        & SAMFeat\cite{wu2025segment} {\scriptsize{(TIP’25)}}                & -- & -- & -- & -- & -- & -- & \textbf{90.2} & \underline{96.0} & \underline{98.5} & \underline{75.9} & \textbf{89.5} & 95.8 \\
        & \textbf{GESS(Ours)}                                              & \underline{86.4} & \textbf{94.2} & \underline{97.2} & \underline{82.7} & \textbf{93.9} & 98.0 & \underline{89.9} & \textbf{96.1} & \textbf{98.9} & \textbf{76.4} & 87.4 & 95.3 \\
        \bottomrule
        \end{tabular}
        \label{Visual_Localization}
    \end{table*}

\subsubsection{Dataset}
The Aachen Day–Night dataset\cite{aachen} is a benchmark for city-scale visual localization, focusing on appearance variations under day-night conditions, and is available in two versions. To ensure a rigorous and fair comparison, we strictly adhere to the standardized evaluation protocol\footnote{https://www.visuallocalization.net} and utilize the established visual localization pipeline\footnote{https://github.com/GrumpyZhou/image-matching-toolbox} for model assessment. Following the SAMFeat\cite{wu2025segment} protocol, we cap the maximum number of keypoints per image at $10,000$.

\begin{table*}[htbp]
    \setlength{\tabcolsep}{14pt}
    \renewcommand{\arraystretch}{1.3}
    \centering
    \caption{Pose estimation results on the MegaDepth1500 and ScanNet datasets.}
    \label{tab:megadepth1500_modified}
    \begin{threeparttable}
        \begin{tabular}{l ccc ccc}
        \toprule
        \multirow{2.5}{*}{\textbf{Methods}} 
        & \multicolumn{3}{c}{\textbf{MegaDepth-1500 Benchmark}} 
        & \multicolumn{3}{c}{\textbf{ScanNet Benchmark}}\\
        \cmidrule(lr){2-4} \cmidrule(lr){5-7}
        & @5$^\circ$$\uparrow$ & @10$^\circ$$\uparrow$ & @20$^\circ$$\uparrow$ 
        & @5$^\circ$$\uparrow$ & @10$^\circ$$\uparrow$ & @20$^\circ$$\uparrow$\\
        \midrule
        SuperPoint~\cite{detone2018superpoint} {\scriptsize{(CVPRW’18)}}                       & 40.8 & 54.3 & 65.3 & 12.8 & 25.0 & 37.7 \\
        D2-Net~\cite{d2net} {\scriptsize{(CVPR’19)}}                                           & 35.4 & 49.5 & 61.9 & 12.8 & 25.2 & 39.0 \\
        R2D2~\cite{r2d2} {\scriptsize{(NeurIPS’19)}}                                           & 48.5 & 63.4 & 74.7 & 11.9 & 23.3 & 35.4 \\
        DISK~\cite{disk} {\scriptsize{(NeurIPS’20)}}                                           & 48.4 & 60.9 & 70.8 & 7.9  & 16.9 & 27.0 \\  
        ALIKE-N \cite{zhao2022alike} {\scriptsize{(TMM’22)}}                                   & 44.6 & 58.3 & 69.3 & 5.9  & 11.9 & 18.7 \\
        AWDesc-CA(2K) \cite{wang2023attention} {\scriptsize{(TPAMI’23)}}                       & 50.3 & 63.9 & 74.1 & 14.2 & 27.2 & 40.9 \\
        SFD2 \cite{xue2023sfd2} {\scriptsize{(CVPR’23)}}                                       & 45.0 & 58.3 & 67.9 & 12.1 & 23.9 & 36.2 \\              
        XFeat \cite{potje2024cvpr} {\scriptsize{(CVPR’24)}}                                    & 35.6 & 50.3 & 63.4 & \textbf{15.8} & \textbf{30.5} & \textbf{45.3} \\   
        LiftFeat \cite{liu2025liftfeat} {\scriptsize{(ICRA’25)}}                               & 46.3 & 60.1 & 71.4 & 15.1 & \underline{29.6} & \underline{44.5} \\
        SAMFeat \cite{wu2025segment} {\scriptsize{(TIP’25)}}                                   & \underline{52.3} & \underline{66.2} & \underline{76.4} & \underline{15.2} & 27.9 & 41.4 \\
        \textbf{GESS(Ours)}                                                                   & \textbf{53.9} & \textbf{67.0} & \textbf{76.7} & \textbf{15.8} & 29.3 & 42.9 \\
        \bottomrule
        \end{tabular}
    \end{threeparttable}
\end{table*}

\subsubsection{Metrics}
Evaluation is performed using the standard pose error thresholds of $(2^{\circ}, 0.25)$, $(5^{\circ}, 0.5)$, and $(10^{\circ}, 5.0)$. We categorize the baseline methods into two groups: Group ``V'', which is specifically optimized for localization datasets, and Group ``G'', which is designed for general-purpose vision tasks.

\subsubsection{Results}
Our model achieves superior performance in this task, with the quantitative results summarized in the Tab. \ref{Visual_Localization}. In the Aachen V1.1\cite{aachen} day scenarios, mid-to-high precision metrics reach $96.1\%$ and $98.9\%$, respectively, ranking first in Group G. Notably, in the most challenging high-precision night tasks, our method attains a score of $76.4\%$, significantly outperforming SOTA algorithms such as SAMFeat\cite{wu2025segment}. This performance advantage is primarily attributed to the UTCF module, which enhances descriptor discriminability through the deep fusion of multi-source cues, and the SDAK mechanism, which filters out unstable keypoints via geometric analysis. The slight discrepancies observed in certain coarse-grained metrics reflect the ``quality-priority'' filtering strategy of SDAK. By trading off feature recall in low-precision tasks for exceptional geometric stability and semantic reliability under stringent matching conditions, this design establishes a decisive advantage in ultra-high-precision localization.

 \begin{figure*}[htbp]
    \centering
    \includegraphics[width=0.95\textwidth]{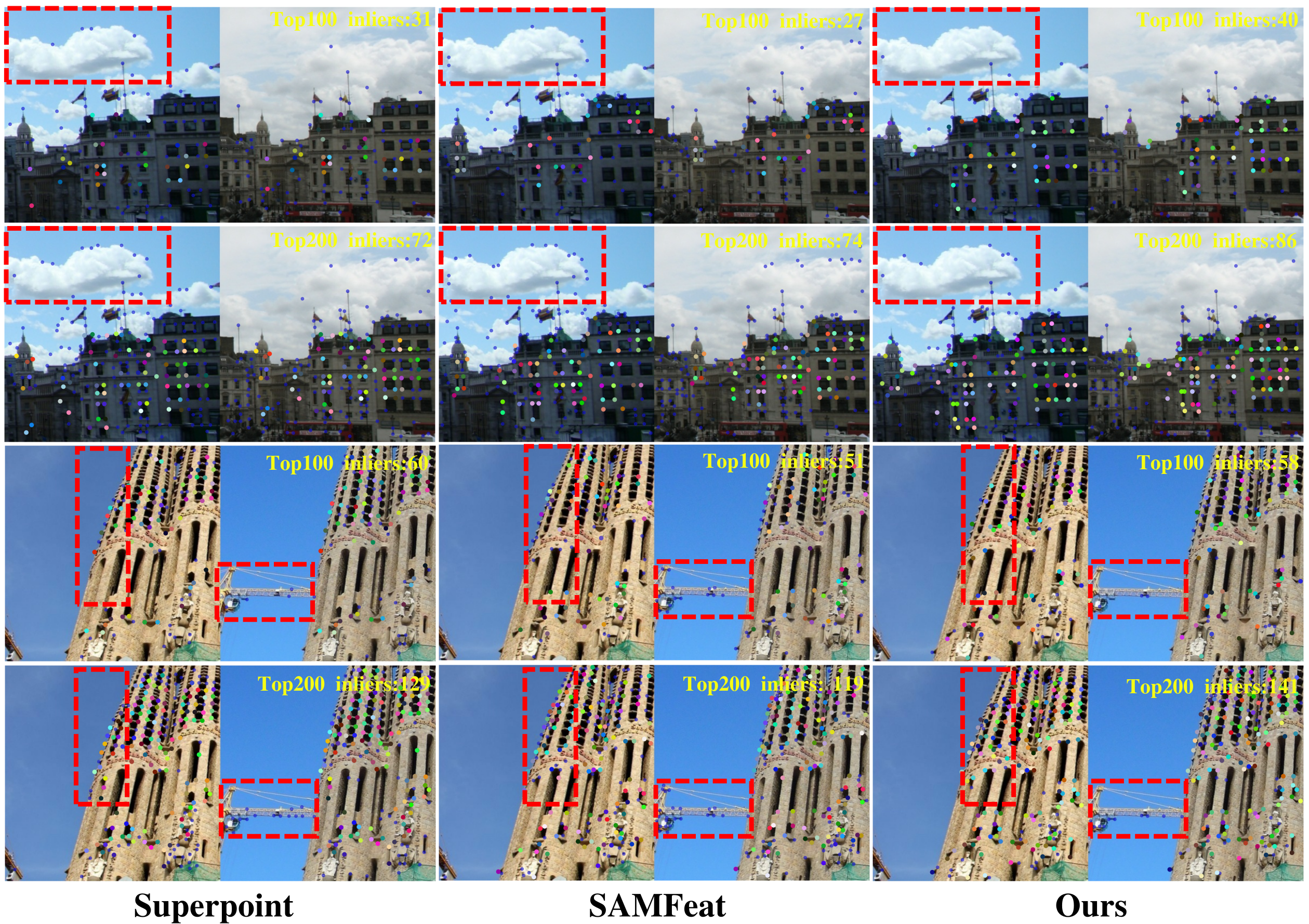}
    \caption{Qualitative comparison of feature stability in complex scenes. Our method effectively suppresses spurious detections in dynamic regions (e.g., clouds) and significantly reduces unstable feature extraction along depth discontinuities.}
    \label{3dandseg}
\end{figure*}

\subsection{Relative pose estimation}

    \begin{table*}[htbp]
        \newcommand{\mc}[1]{\multicolumn{2}{c}{#1}}
        \centering
        \renewcommand{\arraystretch}{1.3}
        \caption{3D reconstruction results on the ETH Dataset.}
        \setlength{\tabcolsep}{5pt}
        \label{3drecon}
        \begin{tabular}{l cc cc cc cc cc cc}
        \toprule
        & \multicolumn{6}{c}{Large-Scale} & \multicolumn{6}{c}{Small-Scale} \\
        \cmidrule(lr){2-7} \cmidrule(lr){8-13}
        Methods & 
        \mc{Madrid metropolis} & 
        \mc{Gendarmenmarkt} & 
        \mc{Tower of London} & 
        \mc{Herzjesu} & 
        \mc{Fountain} & 
        \mc{South building} \\
        \cmidrule(lr){2-7} \cmidrule(lr){8-13}
        & \multicolumn{6}{c}{Track Length ($\uparrow$) / Reproj. Error ($\downarrow$)} 
        & \multicolumn{6}{c}{Track Length ($\uparrow$) / Reproj. Error ($\downarrow$)} \\
        \midrule
            SuperPoint~\cite{detone2018superpoint} {\scriptsize{(CVPRW’18)}} 
            & \mc{5.76 / 1.03} 
            & \mc{4.10 / 1.05} 
            & \mc{4.35 / 0.99} 
            & \mc{4.42 / 0.73} 
            & \mc{4.50 / 0.81} 
            & \mc{5.52 / 0.65} \\
            
            D2-Net~\cite{d2net} {\scriptsize{(CVPR’19)}} 
            & \mc{6.45 / 0.87} 
            & \mc{2.09 / 0.98} 
            & \mc{2.16 / 0.89} 
            & \mc{3.39 / 1.15} 
            & \mc{3.58 / 1.21} 
            & \mc{4.74 / 1.13} \\
            
            R2D2~\cite{r2d2} {\scriptsize{(NeurIPS’19)}} 
            & \mc{2.09 / \underline{0.83}} 
            & \mc{4.46 / 1.11} 
            & \mc{2.27 / 0.98} 
            & \mc{\textbf{5.24} / 1.15} 
            & \mc{\textbf{6.11} / 0.94} 
            & \mc{5.26 / 0.95} \\
            
            ALIKE-N~\cite{zhao2022alike} {\scriptsize{(TMM’22)}} 
            & \mc{7.48 / 0.99} 
            & \mc{6.46 / 1.01} 
            & \mc{3.49 / 0.95} 
            & \mc{3.83 / 0.82} 
            & \mc{4.62 / 0.79} 
            & \mc{\textbf{6.31} / \underline{0.53}} \\
            
            SFD2~\cite{xue2023sfd2} {\scriptsize{(CVPR’23)}} 
            & \mc{4.74 / 1.02} 
            & \mc{5.20 / 1.09} 
            & \mc{4.54 / 1.02} 
            & \mc{4.39 / 0.69} 
            & \mc{4.54 / \underline{0.62}} 
            & \mc{\underline{6.21} / 0.60} \\
            
            AWDesc-CA(2K) \cite{wang2023attention} {\scriptsize{(TPAMI’23)}} 
            & \mc{8.14 / 0.94} 
            & \mc{\underline{7.86} / 1.02} 
            & \mc{7.61 / 0.90} 
            & \mc{4.33 / 0.70} 
            & \mc{4.67 / 0.71} 
            & \mc{5.01 / 0.60} \\
            
            SAMFeat~\cite{wu2025segment} {\scriptsize{(TIP’25)}} 
            & \mc{\underline{8.67} / 0.92} 
            & \mc{7.61 / \underline{0.95}} 
            & \mc{\underline{7.76} / \underline{0.83}} 
            & \mc{4.02 / \underline{0.65}} 
            & \mc{4.22 / 0.65} 
            & \mc{4.97 / 0.58} \\
            
            \textbf{GESS(Ours)} 
            & \mc{\textbf{8.93} / \textbf{0.77}} 
            & \mc{\textbf{8.12} / \textbf{0.84}} 
            & \mc{\textbf{8.56} / \textbf{0.69}} 
            & \mc{\underline{4.49} / \textbf{0.60}} 
            & \mc{\underline{4.81} / \textbf{0.56}} 
            & \mc{5.24 / \textbf{0.47}} \\
        \bottomrule
        \end{tabular}
    \end{table*}

\subsubsection{Dataset}
MegaDepth-1500 \cite{li2018megadepth} is a large-scale outdoor scene dataset comprising diverse landmarks, widely utilized for evaluating feature matching and relative pose estimation algorithms. ScanNet \cite{dai2017scannet} is a comprehensive indoor RGB-D dataset featuring significant viewpoint and illumination variations, frequently serving as a benchmark for indoor camera pose estimation.

\subsubsection{Metrics}
Following the evaluation protocol established in \cite{potje2024cvpr, lindenberger_2023_lightglue}, we adopt the Area Under the Curve (AUC) of the pose error at thresholds of 5°, 10°, and 20° as our evaluation metrics.

\subsubsection{Results}
As shown in the Tab. \ref{tab:megadepth1500_modified}, our method performs excellently on the {MegaDepth-1500}\cite{li2018megadepth} outdoor benchmark, achieving a pose accuracy of $53.9$ at $5^{\circ}$. Furthermore, our approach exhibits robust generalization capabilities in cross-domain ScanNet\cite{dai2017scannet} indoor evaluations. It is worth noting that while XFeat\cite{potje2024cvpr} and LiftFeat\cite{liu2025liftfeat} show strong performance on these metrics, their success is primarily attributed to their training sets, which incorporate the large-scale COCO\cite{lin2014coco} dataset alongside MegaDepth\cite{megadepth}, providing them with highly diverse multi-domain feature representations. In contrast, our method follows the same training protocol as SAMFeat\cite{wu2025segment}. Despite the absence of indoor scenes or general object data during training, our model consistently outperforms it, proving that our method significantly enhances feature universality and effectively captures shared cross-scene characteristics even in data-constrained scenarios.

\subsection{3D Reconstruction}

\subsubsection{Datasets}
The ETH Local Feature Benchmark \cite{schonberger2017comparative} serves as an authoritative framework for evaluating the robustness and geometric accuracy of local features in 3D reconstruction tasks. By employing the COLMAP \cite{schonberger2016structure} incremental Structure-from-Motion and MultiView Stereo pipeline, this benchmark directly maps low-level feature matching quality to high-level reconstruction performance. To validate generalization capability, we conducted a comprehensive evaluation across six representative scenarios of varying scales, ensuring the reliability of our results across diverse scene complexities.

\subsubsection{Metrics}
Following the evaluation protocols of PosFeat\cite{li2022decoupling} and SAMFeat\cite{wu2025segment}, we adopt Mean Reprojection Error and Mean Track Length as our primary metrics to evaluate the geometric localization accuracy of 3D reconstructions and the stability of multi-view matching, respectively.

\subsubsection{Results}
As illustrated in the Tab. \ref{3drecon}, our method demonstrates superior performance in this task. This effectively proves that by incorporating surface normal and depth stability cues, the model can establish more stringent geometric constraints and achieve sub-pixel localization accuracy.

Feature stability visualization (Fig. \ref{3dandseg}) further elucidates the geometric precision advantages of our method. The results demonstrate that the SDAK mechanism not only effectively filters dynamic interferences, such as clouds, but also significantly suppresses unstable keypoint detections caused by geometric abruptness, particularly at depth boundaries. By jointly mitigating dynamic noise and geometric instabilities, our approach ensures high spatial reliability for feature points while substantially enhancing reconstruction robustness in complex environments. This qualitative stability ultimately translates into superior track consistency and minimal reprojection errors in our quantitative assessments.

\subsection{Ablation Studies}
\subsubsection{Effectiveness of Individual Components}
As shown in the Tab. \ref{tab:ablation_hpatches_v4}, we establish the baseline using MTLDesc\cite{wang2022mtldesc} with relevant modules removed, first validating the effectiveness of various cues on the HPatches\cite{hpatches}, dataset. Experimental results demonstrate that, built upon the common foundational appearance cues, the introduction of each cue significantly bolsters matching performance. Ultimately, the model achieves peak performance after integrating all proposed cues into our framework. Furthermore, the model maintains an average inference efficiency of \textbf{56 FPS} at a resolution of $480 \times 640$, demonstrating strong potential for practical applications.

Building upon the fine-grained cue verification on HPatches\cite{hpatches}, we consolidate the components into three functional modules—MLSNet, UTCF, and SDAK—to assess their system-level contributions to downstream pose estimation. As illustrated in the Tab. \ref{tab:pose_ablation_final}, the estimation accuracy demonstrates a consistent upward trend as these core modules are incrementally integrated. Specifically, MLSNet substantially boosts the pose estimation accuracy in outdoor scenarios, underscoring the critical role of multi-scale features in handling scale consistency. UTCF effectively bolsters descriptor discriminativeness and matching robustness by leveraging the coupling of semantic and normal cues. Finally, SDAK further bolsters localization robustness and matching capability by filtering out unstable keypoints in regions with depth discontinuities, enabling the model to achieve optimal performance across all evaluation metrics.

\begin{figure*}[htbp]
    \centering
    \includegraphics[width=1\textwidth]{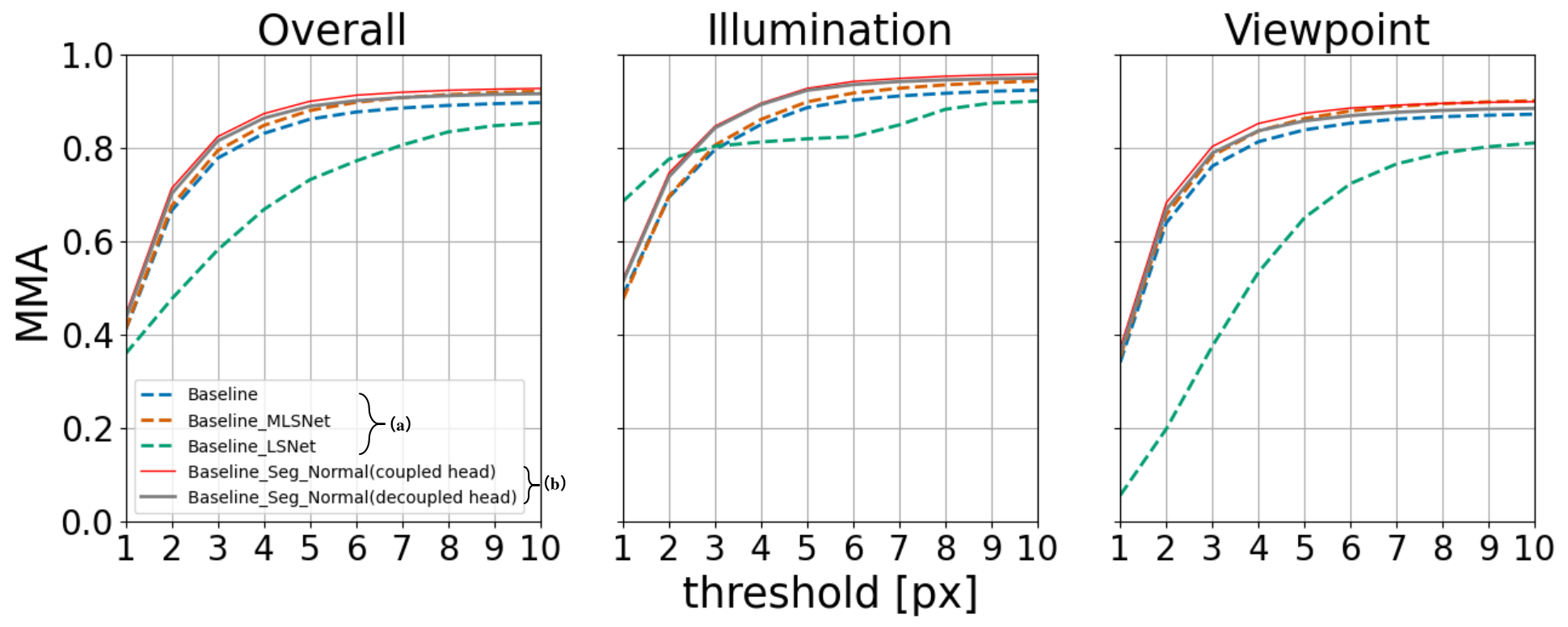}
    \caption{Ablation studies on HPatches. (a) Impact of multi-scale backbone configurations. (b) Effect of the Semantic-Normal Head components.}
    \label{fig:Ablation}
\end{figure*}

\begin{figure}[htbp]
    \centering
    \includegraphics[width=0.45\textwidth]{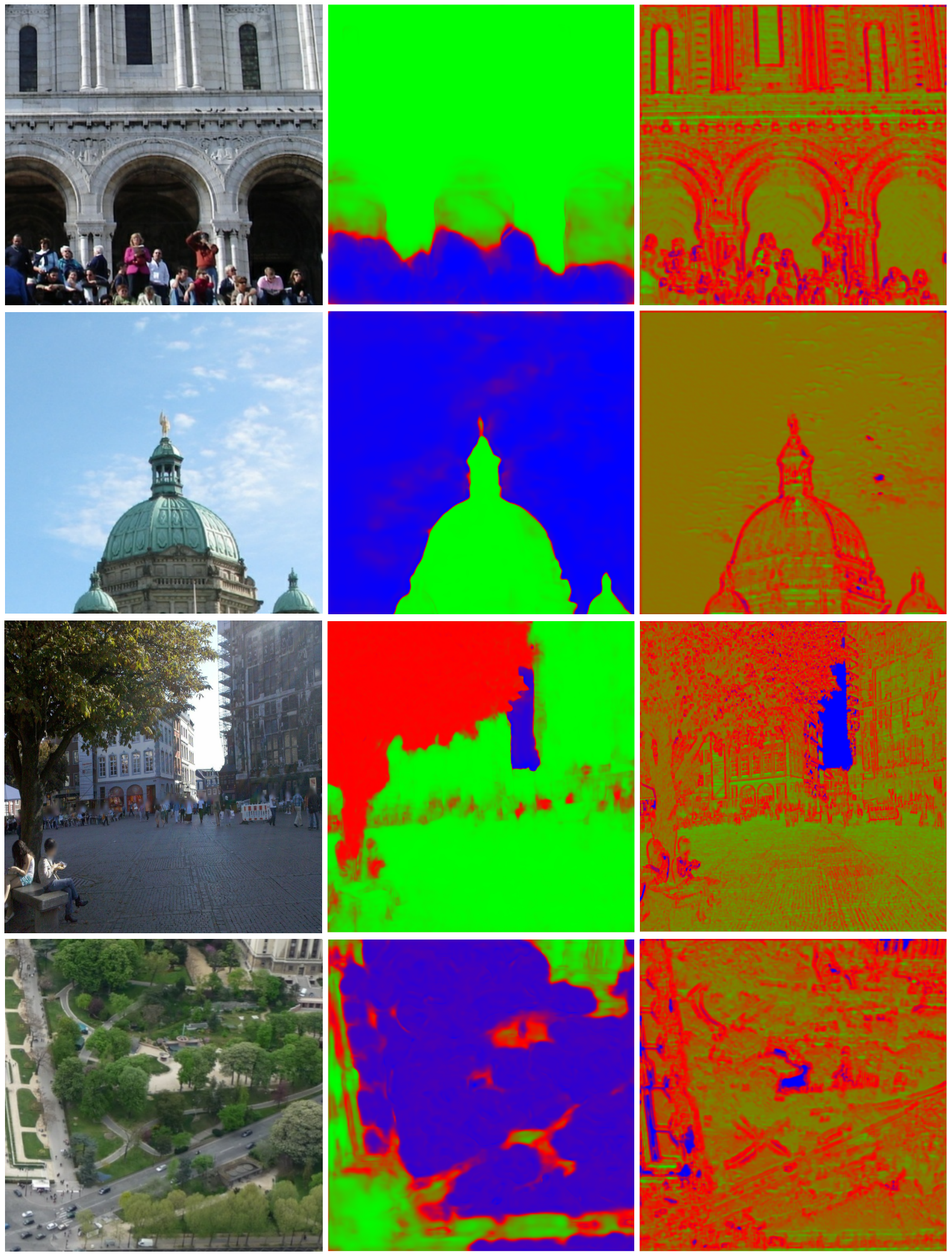}
    \caption{Qualitative visualization of semantic reliability and geometric stability. From left to right: original images, semantic reliability maps (anchoring static structures), and geometric stability heatmaps (perceiving geometric boundaries).}
    \label{fig:heatmap}
\end{figure}

\begin{table}[htbp]
    \centering
    \begin{threeparttable}
        \setlength{\tabcolsep}{2pt} 
        \renewcommand{\arraystretch}{1.3}
        \caption{Ablation experiment results on the HPatches dataset.}
        \label{tab:ablation_hpatches_v4}
        
        \begin{tabular}{cc ccc ccc}
            \toprule
            \multicolumn{5}{c}{Multi-Cues} & \multicolumn{3}{c}{Metrics} \\
            \cmidrule(lr){1-5} \cmidrule(lr){6-8}
            \multicolumn{2}{c}{Appearance} & \multirow{2}{*}[-0.5ex]{Semantic} & \multirow{2}{*}[-0.5ex]{Normal} & \multirow{2}{*}[-0.5ex]{Stability} & \multirow{2}{*}[-0.5ex]{MMA@3} & \multirow{2}{*}[-0.5ex]{AUC@2} & \multirow{2}{*}[-0.5ex]{AUC@5} \\
            \cmidrule(lr){1-2}
            $\text{M}_1$\tnote{*} & $\text{M}_2$\tnote{\dag} & & & & & & \\
            \midrule
            \checkmark &            &            &            &            & 0.778 & 0.537 & 0.727 \\ 
            \checkmark &            & \checkmark &            &            & 0.823 & 0.569 & 0.767 \\
            \checkmark &            &            & \checkmark &            & 0.815 & 0.572 & 0.762 \\
            \checkmark &            &            &            & \checkmark & 0.788 & 0.544 & 0.736 \\
            \checkmark &            & \checkmark & \checkmark &            & 0.824 & 0.577 & 0.770 \\
           & \checkmark &            &            &            & 0.794 & 0.543 & 0.740 \\
           & \checkmark & \checkmark &            &            & 0.823 & 0.570 & 0.768 \\
           & \checkmark &            & \checkmark &            & 0.822 & 0.569 & 0.768 \\
           & \checkmark &            &            & \checkmark & 0.823 & 0.572 & 0.769 \\
           & \checkmark & \checkmark & \checkmark &            & 0.826 & 0.568 & 0.770 \\
           & \checkmark & \checkmark & \checkmark & \checkmark & \textbf{0.835} & \textbf{0.576} & \textbf{0.778} \\
            \bottomrule
        \end{tabular}
        
        \begin{tablenotes}
            \footnotesize
            \item[*] $M_1$: The backbone of MTLDesc \cite{wang2022mtldesc}.
            \item[\dag] $M_2$: MLSNet.
        \end{tablenotes}
    \end{threeparttable}
\end{table}

To qualitatively interpret the underlying mechanism behind the aforementioned performance improvements, the Fig. \ref{fig:heatmap} visualizes the semantic reliability and geometric stability heatmaps in complex scenarios. The semantic reliability maps demonstrate the model's ability to accurately anchor on long-term static structures, such as buildings, while effectively suppressing dynamic distractors like clouds and pedestrians. Simultaneously, the geometric stability maps guide the feature extraction process to avoid geometrically fragile regions by perceiving depth discontinuities. This screening process, driven by the dual-constraint mechanism of ``semantic filtering and geometric awareness'', essentially enhances the signal-to-noise ratio and environmental invariance of the feature descriptors by explicitly suppressing non-coherent noise in complex backgrounds, thereby providing a solid foundation for the extraction of robust and highly discriminative features.

\begin{table}[H]
  \centering
  \caption{Ablation study of pose estimation \\ on MegaDepth-1500\cite{li2018megadepth} and ScanNet\cite{dai2017scannet} datasets.}
  \setlength{\tabcolsep}{3.5pt}
  \begin{tabular}{l ccc @{\hspace{15pt}} ccc}
    \toprule
    \multirow{2}{*}{Methods} & \multicolumn{3}{c}{\textbf{MegaDepth-1500}} & \multicolumn{3}{c}{\textbf{ScanNet}} \\
    \cmidrule(lr){2-4} \cmidrule(lr){5-7}
    & @$5^{\circ}\uparrow$ & @$10^{\circ}\uparrow$ & @$20^{\circ}\uparrow$ & @$5^{\circ}\uparrow$ & @$10^{\circ}\uparrow$ & @$20^{\circ}\uparrow$ \\
    \midrule
    Baseline              & 48.3 & 61.7 & 71.8 & 12.79 & 25.33 & 38.53 \\
    + MLSNet              & 52.0 & 65.4 & 75.5 & 14.00 & 26.66 & 39.84 \\
    + UTCF                & 53.1 & 66.4 & 76.5 & 14.30 & 27.26 & 40.22 \\
    \textbf{Full (SDAK)}  & \textbf{53.9} & \textbf{67.0} & \textbf{76.7} & \textbf{15.81} & \textbf{29.25} & \textbf{42.87} \\
    \bottomrule
  \end{tabular}
  \label{tab:pose_ablation_final}
\end{table}

\subsubsection{Ablation on Multi-Scale Backbone Architecture}
To further validate the contribution of multi-scale features to matching performance, we replaced the backbone with LSNet\cite{wang2025lsnetlargefocussmall}, which relies solely on deep-layer single-scale features (as indicated by the green curve in Fig. \ref{fig:Ablation}), based on the baseline of the aforementioned ablation study. The experimental results demonstrate that, compared to MLSNet with its multi-scale architecture, the MMA metrics for LSNet experience a significant decline. This outcome profoundly reflects the pivotal role of multi-scale information in the field of local feature learning: although deep-layer features possess robust macro-level discriminative power, they inevitably lose substantial critical texture details and spatial resolution during successive downsampling operations. Consequently, it becomes challenging to preserve sub-pixel localization precision under complex viewpoint variations.

\subsubsection{Ablation on Semantic-Normal Head Architecture}
To further validate the effectiveness of the Semantic-Normal Head design, we conduct ablation experiments on two distinct head architectures: the coupled head (red solid line) and the decoupled head (gray solid line). As shown in Fig. \ref{fig:Ablation}), the coupled head marginally outperforms the decoupled head across the three test subsets. This result suggests that feature sharing between the two tasks facilitates the propagation of task-relevant information, enabling the coupled head to learn more discriminative matching features.

\section{Conclusion}
This paper proposes a multi-cue guided robust local feature learning framework that enhances matching performance in complex scenarios by deeply integrating semantic, surface normal, and depth stability cues. The framework introduces a Semantic-Normal Coupling Prediction Head to establish an organic link between macro-level semantics and micro-level geometric details, thereby effectively mitigating common optimization conflicts in multi-task learning. Meanwhile, the UTCF module is employed to achieve adaptive feature injection, ensuring high discriminativity of the descriptors. Furthermore, in conjunction with a dedicated Depth Stability Prediction Head, the SDAK mechanism dynamically re-weights keypoint responses by evaluating local depth discontinuities and semantic stability. This ensures that high-gain descriptors are anchored to physically meaningful and robustly localized keypoints, significantly enhancing the versatility of the features in complex environments. Extensive experiments validate the effectiveness of this multi-cue collaborative paradigm. Future work will focus on on-device real-time deployment and multi-modal information fusion to further advance the model's generalization and practical utility in complex real-world settings.


\bibliographystyle{IEEEtran}
\bibliography{references}

@inproceedings{d2net,
  address    = {Long Beach, CA, USA},
  title      = {D2-{Net}: {A} {Trainable} {CNN} for {Joint} {Description} and {Detection} of {Local} {Features}},
  isbn       = {978-1-72813-293-8},
  shorttitle = {D2-{Net}},
  language   = {en},
  booktitle  = {2019 {IEEE}/{CVF} {Conference} on {Computer} {Vision} and {Pattern} {Recognition}},
  author     = {Dusmanu, Mihai and Rocco, Ignacio and Pajdla, Tomas and Pollefeys, Marc and Sivic, Josef and Torii, Akihiko and Sattler, Torsten},
  month      = jun,
  year       = {2019},
  pages      = {8084--8093}
}

@inproceedings{li2018megadepth,
  title={Megadepth: Learning single-view depth prediction from internet photos},
  author={Li, Zhengqi and Snavely, Noah},
  booktitle={Proceedings of the IEEE conference on computer vision and pattern recognition},
  pages={2041--2050},
  year={2018}
}

@inproceedings{dai2017scannet,
  title={Scannet: Richly-annotated 3d reconstructions of indoor scenes},
  author={Dai, Angela and Chang, Angel X and Savva, Manolis and Halber, Maciej and Funkhouser, Thomas and Nie{\ss}ner, Matthias},
  booktitle={Proceedings of the IEEE conference on computer vision and pattern recognition},
  pages={5828--5839},
  year={2017}
}

@inproceedings{schonberger2016structure,
  title={Structure-from-motion revisited},
  author={Schonberger, Johannes L and Frahm, Jan-Michael},
  booktitle={Proceedings of the IEEE conference on computer vision and pattern recognition},
  pages={4104--4113},
  year={2016}
}

@inproceedings{schonberger2017comparative,
  title={Comparative evaluation of hand-crafted and learned local features},
  author={Schonberger, Johannes L and Hardmeier, Hans and Sattler, Torsten and Pollefeys, Marc},
  booktitle={Proceedings of the IEEE conference on computer vision and pattern recognition},
  pages={1482--1491},
  year={2017}
}

@article{fan2022learning,
  title={Learning semantic-aware local features for long term visual localization},
  author={Fan, Bin and Zhou, Junjie and Feng, Wensen and Pu, Huayan and Yang, Yuzhu and Kong, Qingqun and Wu, Fuchao and Liu, Hongmin},
  journal={IEEE Transactions on Image Processing},
  volume={31},
  pages={4842--4855},
  year={2022},
  publisher={IEEE}
}

@article{lowe2004distinctive,
  title={Distinctive image features from scale-invariant keypoints},
  author={Lowe, David G},
  journal={International journal of computer vision},
  volume={60},
  number={2},
  pages={91--110},
  year={2004},
  publisher={Springer}
}

@inproceedings{luo2020aslfeat,
  title={Aslfeat: Learning local features of accurate shape and localization},
  author={Luo, Zixin and Zhou, Lei and Bai, Xuyang and Chen, Hongkai and Zhang, Jiahui and Yao, Yao and Li, Shiwei and Fang, Tian and Quan, Long},
  booktitle={Proceedings of the IEEE/CVF conference on computer vision and pattern recognition},
  pages={6589--6598},
  year={2020}
}

@inproceedings{xue2022efficient,
  title={Efficient large-scale localization by global instance recognition},
  author={Xue, Fei and Budvytis, Ignas and Reino, Daniel Olmeda and Cipolla, Roberto},
  booktitle={Proceedings of the IEEE/CVF Conference on Computer Vision and Pattern Recognition},
  pages={17348--17357},
  year={2022}
}

@InProceedings{lindenberger_2023_lightglue,
  title     = {{LightGlue: Local Feature Matching at Light Speed}},
  author    = {Philipp Lindenberger and
               Paul-Edouard Sarlin and
               Marc Pollefeys},
  booktitle = {International Conference on Computer Vision (ICCV)},
  year      = {2023}
}

@inproceedings{r2d2,
  title     = {{R2D2}: {Repeatable} and {Reliable} {Detector} and {Descriptor}},
  language  = {en},
  booktitle = {{NeurIPS}},
  author    = {Revaud, Jerome and Weinzaepfel, Philippe and Souza, C{\'e}sar De and Pion, Noe and Csurka, Gabriela and Cabon, Yohann and Humenberger, Martin},
  year      = {2019},
  pages     = {12}
}

@misc{yi2016liftlearnedinvariantfeature,
      title={LIFT: Learned Invariant Feature Transform}, 
      author={Kwang Moo Yi and Eduard Trulls and Vincent Lepetit and Pascal Fua},
      year={2016},
      eprint={1603.09114},
      archivePrefix={arXiv},
      primaryClass={cs.CV}
}

@article{liu2025liftfeat,
  title={LiftFeat: 3D Geometry-Aware Local Feature Matching},
  author={Liu, Yepeng and Lai, Wenpeng and Zhao, Zhou and Xiong, Yuxuan and Zhu, Jinchi and Cheng, Jun and Xu, Yongchao},
  journal={arXiv preprint arXiv:2505.03422},
  year={2025}
}

@INPROCEEDINGS{potje2024cvpr,
  author={Potje, Guilherme and Cadar, Felipe and Araujo, André and Martins, Renato and Nascimento, Erickson R.},
  booktitle={2024 IEEE/CVF Conference on Computer Vision and Pattern Recognition (CVPR)}, 
  title={XFeat: Accelerated Features for Lightweight Image Matching}, 
  year={2024},
  pages={2682-2691},
  doi={10.1109/CVPR52733.2024.00259}}

@inproceedings{superpoint,
  title     = {{SuperPoint}: {Self}-{Supervised} {Interest} {Point} {Detection} and {Description}},
  language  = {en},
  booktitle = {Proceedings of the {IEEE} {Conference} on {Computer} {Vision} and {Pattern} {Recognition} {Workshops}},
  author    = {DeTone, Daniel and Malisiewicz, Tomasz and Rabinovich, Andrew},
  year      = {2018},
  pages     = {224--236}
}

@ARTICLE{4674368,
  author={Rosten, Edward and Porter, Reid and Drummond, Tom},
  journal={IEEE Transactions on Pattern Analysis and Machine Intelligence}, 
  title={Faster and Better: A Machine Learning Approach to Corner Detection}, 
  year={2010},
  volume={32},
  number={1},
  pages={105-119},
  doi={10.1109/TPAMI.2008.275}}

@InProceedings{10.1007/978-3-642-15561-1_56,
author="Calonder, Michael
and Lepetit, Vincent
and Strecha, Christoph
and Fua, Pascal",
editor="Daniilidis, Kostas
and Maragos, Petros
and Paragios, Nikos",
title="BRIEF: Binary Robust Independent Elementary Features",
booktitle="Computer Vision -- ECCV 2010",
year="2010",
publisher="Springer Berlin Heidelberg",
address="Berlin, Heidelberg",
pages="778--792",
isbn="978-3-642-15561-1"
}

@inproceedings{aslfeat,
  title      = {{ASLFeat}: {Learning} {Local} {Features} of {Accurate} {Shape} and {Localization}},
  shorttitle = {{ASLFeat}},
  booktitle  = {Proceedings of the IEEE/CVF Conference on Computer Vision and Pattern Recognition},
  author     = {Luo, Zixin and Zhou, Lei and Bai, Xuyang and Chen, Hongkai and Zhang, Jiahui and Yao, Yao and Li, Shiwei and Fang, Tian and Quan, Long},
  month      = apr,
  year       = {2020}
}

@inproceedings{disk,
  title      = {{DISK}: {Learning} local features with policy gradient},
  shorttitle = {{DISK}},
  language   = {en},
  booktitle  = {Neural {IPS}},
  author     = {Tyszkiewicz, Micha{\l } J. and Fua, Pascal and Trulls, Eduard},
  month      = jun,
  year       = {2020}
}

@ARTICLE{1008392,
  author={Olson, C.F.},
  journal={IEEE Transactions on Pattern Analysis and Machine Intelligence}, 
  title={Maximum-likelihood image matching}, 
  year={2002},
  volume={24},
  number={6},
  pages={853-857},
  doi={10.1109/TPAMI.2002.1008392}
}

@inproceedings{hpatches,
  title      = {{HPatches}: {A} benchmark and evaluation of handcrafted and learned local descriptors},
  shorttitle = {{HPatches}},
  booktitle  = {Proceedings of the {IEEE} {Conference} on {Computer} {Vision} and {Pattern} {Recognition}},
  author     = {Balntas, Vassileios and Lenc, Karel and Vedaldi, Andrea and Mikolajczyk, Krystian},
  year       = {2017},
  pages      = {5173--5182}
}

@ARTICLE{5226635,
  author={Furukawa, Yasutaka and Ponce, Jean},
  journal={IEEE Transactions on Pattern Analysis and Machine Intelligence}, 
  title={Accurate, Dense, and Robust Multiview Stereopsis}, 
  year={2010},
  volume={32},
  number={8},
  pages={1362-1376},
  doi={10.1109/TPAMI.2009.161}
}

@INPROCEEDINGS{8100132,
  author={Tian, Yurun and Fan, Bin and Wu, Fuchao},
  booktitle={2017 IEEE Conference on Computer Vision and Pattern Recognition (CVPR)}, 
  title={L2-Net: Deep Learning of Discriminative Patch Descriptor in Euclidean Space}, 
  year={2017},
  volume={},
  number={},
  pages={6128-6136},
  doi={10.1109/CVPR.2017.649}}

@article{MurArtal2017,
   title={ORB-SLAM2: An Open-Source SLAM System for Monocular, Stereo, and RGB-D Cameras},
   volume={33},
   ISSN={1941-0468},
   DOI={10.1109/tro.2017.2705103},
   number={5},
   journal={IEEE Transactions on Robotics},
   publisher={Institute of Electrical and Electronics Engineers (IEEE)},
   author={Mur-Artal, Raul and Tardos, Juan D.},
   year={2017},
   month=oct, pages={1255–1262}
}

@misc{semantic2018,
      title={Semantic Visual Localization}, 
      author={Johannes L. Schönberger and Marc Pollefeys and Andreas Geiger and Torsten Sattler},
      year={2018},
      eprint={1712.05773},
      archivePrefix={arXiv},
      primaryClass={cs.CV}
}

@ARTICLE{11373611,
  author={Zeng, Longjian and Zhu, Zunjie and Lu, Ming and Zheng, Bolun and Lu, Rongfeng and Wang, Tingyu and Zheng, Zhongtian and Sun, Yaoqi and Yan, Chenggang},
  journal={IEEE Transactions on Circuits and Systems for Video Technology}, 
  title={LLFeat: Noise-Aware Feature Matching under Various Low-Light Conditions}, 
  year={2026},
  volume={},
  number={},
  pages={1-1},
  doi={10.1109/TCSVT.2026.3662204}}

@ARTICLE{10851359,
  author={Wang, Changhao and Zhang, Guanwen and Cheng, Zhengyun and Zhou, Wei},
  journal={IEEE Transactions on Circuits and Systems for Video Technology}, 
  title={KPDepth-VO: Self-Supervised Learning of Scale-Consistent Visual Odometry and Depth With Keypoint Features From Monocular Video}, 
  year={2025},
  volume={35},
  number={6},
  pages={5762-5775},
  doi={10.1109/TCSVT.2025.3533256}}

@InProceedings{lin2014coco,
author="Lin, Tsung-Yi
and Maire, Michael
and Belongie, Serge
and Hays, James
and Perona, Pietro
and Ramanan, Deva
and Doll{\'a}r, Piotr
and Zitnick, C. Lawrence",
editor="Fleet, David
and Pajdla, Tomas
and Schiele, Bernt
and Tuytelaars, Tinne",
title="Microsoft COCO: Common Objects in Context",
booktitle="Computer Vision -- ECCV 2014",
year="2014",
publisher="Springer International Publishing",
address="Cham",
pages="740--755",
}

@article{wang2025moge,
  title={Moge-2: Accurate monocular geometry with metric scale and sharp details},
  author={Wang, Ruicheng and Xu, Sicheng and Dong, Yue and Deng, Yu and Xiang, Jianfeng and Lv, Zelong and Sun, Guangzhong and Tong, Xin and Yang, Jiaolong},
  journal={arXiv preprint arXiv:2507.02546},
  year={2025}
}

@inproceedings{xu2023open,
  title={Open-vocabulary panoptic segmentation with text-to-image diffusion models},
  author={Xu, Jiarui and Liu, Sifei and Vahdat, Arash and Byeon, Wonmin and Wang, Xiaolong and De Mello, Shalini},
  booktitle={Proceedings of the IEEE/CVF conference on computer vision and pattern recognition},
  pages={2955--2966},
  year={2023}
}

@InProceedings{wang2025lsnetlargefocussmall,
    author    = {Wang, Ao and Chen, Hui and Lin, Zijia and Han, Jungong and Ding, Guiguang},
    title     = {LSNet: See Large, Focus Small},
    booktitle = {Proceedings of the IEEE/CVF Conference on Computer Vision and Pattern Recognition (CVPR)},
    month     = {June},
    year      = {2025},
    pages     = {9718-9729}
}

@ARTICLE{8630864,
  author={Ye, Jianming and Zhang, Shiliang and Huang, Tiejun and Rui, Yong},
  journal={IEEE Transactions on Circuits and Systems for Video Technology}, 
  title={CDbin: Compact Discriminative Binary Descriptor Learned With Efficient Neural Network}, 
  year={2020},
  volume={30},
  number={3},
  pages={862-874},
  doi={10.1109/TCSVT.2019.2896095}}

@ARTICLE{8848801,
  author={Liu, Hongmin and Zhang, Qianqian and Fan, Bin and Wang, Zhiheng and Han, Junwei},
  journal={IEEE Transactions on Circuits and Systems for Video Technology}, 
  title={Features Combined Binary Descriptor Based on Voted Ring-Sampling Pattern}, 
  year={2020},
  volume={30},
  number={10},
  pages={3675-3687},
  doi={10.1109/TCSVT.2019.2943595}}

@ARTICLE{10485434,
  author={Li, Dongyue and Du, Songlin},
  journal={IEEE Transactions on Circuits and Systems for Video Technology}, 
  title={ContextMatcher: Detector-Free Feature Matching With Cross-Modality Context}, 
  year={2024},
  volume={34},
  number={9},
  pages={7922-7934},
  doi={10.1109/TCSVT.2024.3383334}}

@article{ernst2002humans,
  title={Humans integrate visual and haptic information in a statistically optimal fashion},
  author={Ernst, Marc O and Banks, Martin S},
  journal={Nature},
  volume={415},
  number={6870},
  pages={429--433},
  year={2002},
  publisher={Nature Publishing Group},
  doi={10.1038/415429a}
}

@INPROCEEDINGS{8578879,
  author={Cipolla, Roberto and Gal, Yarin and Kendall, Alex},
  booktitle={2018 IEEE/CVF Conference on Computer Vision and Pattern Recognition}, 
  title={Multi-task Learning Using Uncertainty to Weigh Losses for Scene Geometry and Semantics}, 
  year={2018},
  volume={},
  number={},
  pages={7482-7491},
  doi={10.1109/CVPR.2018.00781}}

@ARTICLE{10144412,
  author={Xu, Rongtao and Wang, Changwei and Xu, Shibiao and Meng, Weiliang and Zhang, Yuyang and Fan, Bin and Zhang, Xiaopeng},
  journal={IEEE Transactions on Circuits and Systems for Video Technology}, 
  title={DomainFeat: Learning Local Features With Domain Adaptation}, 
  year={2024},
  volume={34},
  number={1},
  pages={46-59},
  doi={10.1109/TCSVT.2023.3282956}}

@article{kirillov2023segany,
  title={Segment Anything},
  author={Kirillov, Alexander and Mintun, Eric and Ravi, Nikhila and Mao, Hanzi and Rolland, Chloe and Gustafson, Laura and Xiao, Tete and Whitehead, Spencer and Berg, Alexander C. and Lo, Wan-Yen and Doll{\'a}r, Piotr and Girshick, Ross},
  journal={arXiv:2304.02643},
  year={2023}
}

@ARTICLE{10741542,
  author={Wang, Li and Zhang, Yunzhou and Ge, Fawei and Bai, Wenjing and Zhang, Jinpeng and Wang, Yifan},
  journal={IEEE Transactions on Circuits and Systems for Video Technology}, 
  title={Learning Local Features by Jointly Semantic-Guided and Task Rewards}, 
  year={2025},
  volume={35},
  number={3},
  pages={2045-2056},
  doi={10.1109/TCSVT.2024.3490797}}

@ARTICLE{1498756,
  author={Mikolajczyk, K. and Schmid, C.},
  journal={IEEE Transactions on Pattern Analysis and Machine Intelligence}, 
  title={A performance evaluation of local descriptors}, 
  year={2005},
  volume={27},
  number={10},
  pages={1615-1630},
  doi={10.1109/TPAMI.2005.188}}

@inproceedings{li2022decoupling,
  title={Decoupling makes weakly supervised local feature better},
  author={Li, Kunhong and Wang, Longguang and Liu, Li and Ran, Qing and Xu, Kai and Guo, Yulan},
  booktitle={Proceedings of the IEEE/CVF conference on computer vision and pattern recognition},
  pages={15838--15848},
  year={2022}
}

@inproceedings{aachen,
  title     = {Benchmarking 6dof outdoor visual localization in changing conditions},
  author    = {Sattler, Torsten and Maddern, Will and Toft, Carl and Torii, Akihiko and Hammarstrand, Lars and Stenborg, Erik and Safari, Daniel and Okutomi, Masatoshi and Pollefeys, Marc and Sivic, Josef and others},
  booktitle = {Proceedings of the IEEE Conference on Computer Vision and Pattern Recognition},
  pages     = {8601--8610},
  year      = {2018}
}

@inproceedings{megadepth,
  title     = {Megadepth: Learning single-view depth prediction from internet photos},
  author    = {Li, Zhengqi and Snavely, Noah},
  booktitle = {Proceedings of the IEEE Conference on Computer Vision and Pattern Recognition},
  pages     = {2041--2050},
  year      = {2018}
}

@article{MENG2025129567,
title = {SPADesc: Semantic and parallel attention with feature description},
journal = {Neurocomputing},
volume = {625},
pages = {129567},
year = {2025},
issn = {0925-2312},
author = {Haijun Meng and Huimin Lu and Bozhi Ding and Qiangchang Wang}
}

@article{zhao2022alike,
  title={Alike: Accurate and lightweight keypoint detection and descriptor extraction},
  author={Zhao, Xiaoming and Wu, Xingming and Miao, Jinyu and Chen, Weihai and Chen, Peter CY and Li, Zhengguo},
  journal={IEEE Transactions on Multimedia},
  year={2022},
  publisher={IEEE}
}

@ARTICLE{10812856,
  author={Li, Jiapeng and Zhang, Ruonan and Li, Ge and Li, Thomas H.},
  journal={IEEE Transactions on Multimedia}, 
  title={SDE2D: Semantic-Guided Discriminability Enhancement Feature Detector and Descriptor}, 
  year={2025},
  volume={27},
  number={},
  pages={275-286},
  doi={10.1109/TMM.2024.3521748}}

@inproceedings{xue2023sfd2,
  title={SFD2: Semantic-guided Feature Detection and Description},
  author={Xue, Fei and Budvytis, Ignas and Cipolla, Roberto},
  booktitle={Proceedings of the IEEE/CVF Conference on Computer Vision and Pattern Recognition},
  pages={5206--5216},
  year={2023}
}

@article{10.1016/j.neucom.2025.131349,
author = {Mo, Yanhan and Yin, Mengxiao and Li, Guiqing and Liao, Junjie and Liang, Zhijie},
title = {SAGA-Feat: A semantic- and geometry-aware network for sparse local feature learning},
year = {2026},
issue_date = {Nov 2025},
publisher = {Elsevier Science Publishers B. V.},
address = {NLD},
volume = {655},
number = {C},
doi = {10.1016/j.neucom.2025.131349},
journal = {Neurocomput.},
month = jan,
numpages = {17}
}

@inproceedings{wang2022mtldesc,
  title={MTLDesc: Looking Wider to Describe Better},
  author={Wang, Changwei and Xu, Rongtao and Zhang, Yuyang and Xu, Shibiao and Meng, Weiliang and Fan, Bin and Zhang, Xiaopeng},
  booktitle={AAAI},
  year={2022},
  organization={AAAI Press}
}

@article{luo2019contextdesc,
  title={ContextDesc: Local Descriptor Augmentation with Cross-Modality Context},
  author={Luo, Zixin and Shen, Tianwei and Zhou, Lei and Zhang, Jiahui and Yao, Yao and Li, Shiwei and Fang, Tian and Quan, Long},
  journal={Computer Vision and Pattern Recognition (CVPR)},
  year={2019}
}

@misc{barrosolaguna2019keynetkeypointdetectionhandcrafted,
      title={Key.Net: Keypoint Detection by Handcrafted and Learned CNN Filters}, 
      author={Axel Barroso-Laguna and Edgar Riba and Daniel Ponsa and Krystian Mikolajczyk},
      year={2019},
      eprint={1904.00889},
      archivePrefix={arXiv},
      primaryClass={cs.CV}
}

@inproceedings{bay2006surf,
  title={Surf: Speeded up robust features},
  author={Bay, Herbert and Tuytelaars, Tinne and Van Gool, Luc},
  booktitle={European conference on computer vision},
  pages={404--417},
  year={2006},
  organization={Springer}
}

@inproceedings{rublee2011orb,
  title={ORB: An efficient alternative to SIFT or SURF},
  author={Rublee, Ethan and Rabaud, Vincent and Konolige, Kurt and Bradski, Gary},
  booktitle={2011 International conference on computer vision},
  pages={2564--2571},
  year={2011},
  organization={Ieee}
}

@inproceedings{detone2018superpoint,
  title={Superpoint: Self-supervised interest point detection and description},
  author={DeTone, Daniel and Malisiewicz, Tomasz and Rabinovich, Andrew},
  booktitle={Proceedings of the IEEE conference on computer vision and pattern recognition workshops},
  pages={224--236},
  year={2018}
}

@article{wang2023attention,
  title={Attention weighted local descriptors},
  author={Wang, Changwei and Xu, Rongtao and Lu, Ke and Xu, Shibiao and Meng, Weiliang and Zhang, Yuyang and Fan, Bin and Zhang, Xiaopeng},
  journal={IEEE Transactions on Pattern Analysis and Machine Intelligence},
  volume={45},
  number={9},
  pages={10632--10649},
  year={2023},
  publisher={IEEE}
}

@article{wu2025segment,
  title={Segment anything model is a good teacher for local feature learning},
  author={Wu, Jingqian and Xu, Rongtao and Wood-Doughty, Zach and Wang, Changwei and Xu, Shibiao and Lam, Edmund Y},
  journal={IEEE Transactions on Image Processing},
  year={2025},
  publisher={IEEE}
}

\end{document}